\pdfoutput=1
\documentclass[10pt,twocolumn,letterpaper]{article}
\usepackage[pagenumbers]{cvpr}      %

\usepackage{cuted} %

\usepackage{currfile} %

\usepackage{caption} %

\usepackage{iftex}\ifPDFTeX\else\usepackage[T1]{fontenc}\fi
\definecolor{cvprblue}{rgb}{0.21,0.49,0.74}
\usepackage[pagebackref,breaklinks,colorlinks,allcolors=cvprblue]{hyperref}
\usepackage{booktabs,amsmath,amssymb,xspace,xcolor,float}

\newcommand{\passk}{pass@$k$\xspace}
\newcommand{\passone}{pass@1\xspace}
\newcommand{\sdm}{SD3.5-M\xspace}

\title{What Does RL Post-Training Teach Text-to-Image Models?\\ A Skill-Wise Analysis of Flow-GRPO}

\author{Cheng Zhang\\ {\normalsize North Carolina State University}\\ {\small czhang48@ncsu.edu} \and Jiesong Liu\\ {\normalsize Purdue University}\\ {\small liu4833@purdue.edu}}

\begin{document}
\maketitle
\begin{abstract}
Reinforcement learning (RL) post-training has become a standard way to improve text-to-image flow models: Flow-GRPO lifts the GenEval score of \sdm from 0.63 to 0.95, a gain read as the model having \emph{learned} compositional generation. We ask how much of that gain the base model's own samples already contained, by measuring \passk over shared noise seeds, one skill at a time, before and after RL, on prompts never used in training. The answer differs by skill. On four of GenEval's six skills, where 32 base samples already solve almost every prompt, the gain is \emph{sharpening}: one post-trained sample is worth 4 to 10 base samples, and pass@32 barely moves. On spatial relations it is \emph{unlocking}: pass@32 rises from 0.69 to 1.00, 128 base samples do not match one post-trained sample, and prompts the base model never solved in 32 samples are solved 93\% of the time, as often as those it solved. The benchmark average, 31 base samples per post-trained sample, describes no skill. Relations carry 35\% of the gain; of the other 65\%, selecting among 8 base samples with the judge behind the reward recovers nine tenths. We propose an account of when each occurs: within a prompt, a group-relative update re-weights only the outcomes its samples contain, and a flow policy cannot widen the noise that produces them; across prompts, shared parameters let successes on some prompts repair a deficit common to many. The same anatomy applies to a text-rendering checkpoint. The account's predictions are tested with a checkpoint trained without KL, within-prompt diversity, and controlled training that holds object classes out, where sharpening comes first and transfers, and unlocking follows, on practised objects before unseen ones.
\end{abstract}

\section{Introduction}
\label{sec:intro}

\begin{figure*}[t]\centering\includegraphics[width=.92\textwidth]{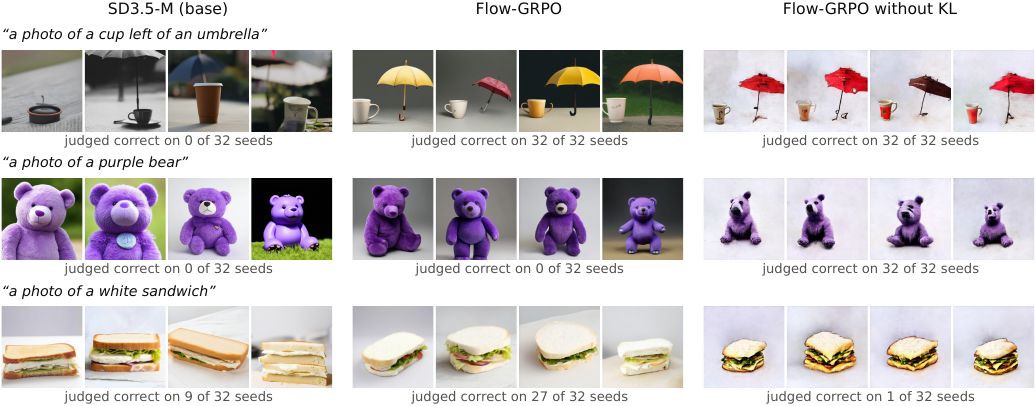}
\caption{\textbf{Three prompts, three models, four shared seeds.} Counts are verdicts of the GenEval judge over 32 seeds. \emph{Top:} a spatial relation the base model never draws in 32 samples is unlocked by RL. \emph{Middle:} the base model reads ``a purple bear'' as a toy, which the judge's detector files under \emph{teddy bear}; Flow-GRPO keeps that reading; without the KL term the model draws the bear the judge expects, in one pale style. \emph{Bottom:} the unregularized model loses a prompt the regularized one had sharpened.}\label{fig:qual}\end{figure*}

A text-to-image model that puts two objects in the spatial relation its prompt asks for in 16\% of its samples does so in 94\% after RL post-training with Flow-GRPO~\cite{flowgrpo} (our measurements on \sdm, \cref{sec:analysis}). The same recipe raises compositional, text-rendering and preference scores~\cite{flowgrpo}, and a family of related methods has followed~\cite{dancegrpo,diffusionnft}. The natural description is that the model \emph{learned} these skills. This paper shows, on Flow-GRPO's public checkpoints, that the description fits some skills and not others, that a single benchmark score cannot tell which, and that a cheap measurement can (\cref{fig:qual}).

The alternative description predates RL for diffusion. For any prompt some noise seeds already land on a correct image; with $k$ attempts and an oracle to pick the best, the success rate is \passk~\cite{codex}. A policy-gradient update on a group of $G$ samples moves probability mass from the failing seeds toward the images the successful seeds produced. If that is all that happens, the post-trained model is the base model with best-of-$G$ selection folded into its weights: strong at \passone, no better at large $k$, and unable to solve a prompt none of its seeds ever solved. For language models this ``sharpening'' reading has been argued both ways~\cite{yue2025does,prorl,nsr}, each side measuring a whole benchmark at once. For text-to-image flow models, whose policy cannot even set the scale of its own noise, the question has not been asked.

We ask it one skill at a time, on prompts never used in training, and both readings turn out to hold (\cref{sec:analysis}). For single objects, two objects, counting and colours, where 32 base samples already solve 347 of 353 GenEval prompts, one sample from the Flow-GRPO checkpoint is worth 4 to 10 base samples and pass@32 barely moves. We call this sharpening. For spatial relations the base model solves 16\% of samples and 69\% of prompts, the post-trained model 94\% and 100\%. The 31 relation prompts the base model never solved in 32 samples are now solved 93\% of the time, as often as those it solved, whereas on the other skills a prompt's rate after RL tracks its rate before; and where the base model draws both objects of such a prompt it arranges them as asked 5\% of the time, the post-trained model 98\%. Selection among 128 base samples does not buy this. We call it unlocking. Attribute binding and text rendering mix the two. Averaged over the benchmark, the base model needs 31 samples to match one post-trained sample, a number that describes no skill: five need 4 to 10, and for relations 128 are not enough.

Why would one training run do two different things? \cref{sec:mechanism} proposes an account and derives predictions that \cref{sec:analysis,sec:method} test. Within a prompt, a group-relative update re-weights only what its group contains, and a flow policy cannot widen its search: the network predicts the mean of a Gaussian whose variance is a fixed hyper-parameter of the ODE-to-SDE conversion~\cite{flowgrpo}. Across prompts, the policy is one shared function. When many prompts fail for the same reason, that reason is a shared direction in parameter space; successes on the solvable prompts push along it, and prompts without successes acquire them. When a failure reflects how the reference model reads one word (a purple bear is a toy), nothing learned elsewhere bears on it and the KL penalty holds it in place; a checkpoint trained without KL releases it (\cref{fig:qual}).

The distinction matters because the two call for different investments. Sharpening can be bought at inference, under the same judge and at $k$ times the sampling cost: a reward good enough to train on is good enough to select with. Unlocking is what training added. Controlled training with 20 held-out object classes (\cref{sec:method}) shows their order: up to 26k samples the gain is worth about one extra draw and coverage barely moves; with 44k to 78k samples spent on relations, prompts the base model never solved begin to be solved, first where both objects were in the training vocabulary and then, more slowly, for held-out objects.

\paragraph{Contributions.} (1) A per-skill \passk anatomy with operational definitions of sharpening and unlocking, two components of an RL gain that the benchmark score conflates, applied to three public Flow-GRPO checkpoints on two tasks and along a LoRA interpolation. (2) An account of both for flow-matching GRPO, from which testable predictions follow. (3) Tests of the predictions: a checkpoint trained without KL, within-prompt diversity, and controlled training with held-out object classes, where early gains are sharpening and unlocking first appears on practised objects.

\section{Background}
\label{sec:background}

\paragraph{Flow matching and SDE conversion.} A rectified-flow model~\cite{sd3,rectifiedflow} learns a velocity field $v_\theta(x_t,t,c)$ that transports noise $x_1$ to data $x_0$ along $dx_t=v_\theta\,dt$; given the seed and the prompt, the image is a deterministic function $x_0=F_\theta(x_1,c)$. Policy gradients need a likelihood, so Flow-GRPO~\cite{flowgrpo} samples with an SDE that has the same marginals, $dx_t=(v_\theta+\tfrac{\sigma_t^2}{2t}(x_t+(1-t)v_\theta))\,dt+\sigma_t\,dw$ with $\sigma_t=a\sqrt{t/(1-t)}$ and a global noise level $a$ (0.7 in the released models). Each denoising step is a Gaussian whose mean depends on the network and whose standard deviation is \emph{fixed}.

\paragraph{GRPO for flow models.} For a prompt $c$, $G$ trajectories are sampled, scored by a reward $r_i$ and assigned advantages $A_i=(r_i-\bar r)/s$ from the group statistics~\cite{grpo,flowgrpo}; a clipped policy-gradient loss on the per-step likelihood ratio and a KL penalty to the reference model complete the objective. The GenEval reward is \emph{graded}: in the released implementation it is the mean of per-clause credits, for the count of each requested object, for its colour, and for its relation to another object, the last being all-or-nothing. Released \sdm checkpoints use $G{=}24$, rank-32 LoRA, 10 denoising steps for collection and 40 at evaluation.

\paragraph{\passk as budgeted coverage.} With $n$ samples of which $m$ are correct, $1-\binom{n-m}{k}/\binom{n}{k}$ is an unbiased estimate of the probability that at least one of $k\le n$ samples is correct~\cite{codex}. We call an image \emph{correct} when the reward above attains its maximum: every clause is met, with exactly the requested object counts. This \emph{strict} rule is binary, which makes \passk well defined. Coverage is always relative to a budget. By \emph{ceiling} we mean pass@$n$ (pass@32 unless stated), and by a \emph{zero-coverage} prompt one the base model never solves in $n{=}32$ samples, which bounds its success probability below 0.09 at 95\% confidence (0.023 for none in 128), not at zero.

\section{Two Routes to a Gain: A Testable Account}
\label{sec:mechanism}

This section gives our account of why one training run does both, and derives predictions that \cref{sec:analysis,sec:method} test. The 553 GenEval prompts of \cref{sec:analysis} are held out from training, so its tests see what reaches unseen prompts.

\paragraph{The noise is exogenous.} In an autoregressive language model the policy's entropy is part of what the network computes, and RL can raise it where uncertainty pays. In flow-matching GRPO the per-step policy is $\mathcal N(\mu_\theta(x_t,t,c),\sigma_t^2 I)$ and the network controls $\mu_\theta$ only; the seed $x_1\sim\mathcal N(0,I)$ and the SDE increments with fixed $\sigma_t$ are not learned, so the policy cannot widen its search on a prompt where every sample fails. What is fixed is the noise, not the variety of the images: the map from noise to image is learned, and an update can fold many seeds onto one image (prediction~4).

\paragraph{Within a prompt: re-weighting.} Consider a group of $G$ samples for a prompt $c$. The gradient raises the likelihood of the trajectories whose reward is above the group mean and lowers that of the others, so a group can teach only what its samples differ in. Count credits are graded; the relation credit is all-or-nothing, so a group in which no sample has the requested arrangement can push toward drawing both objects, not toward arranging them. Because $\mu_\theta$ is smooth across seeds, raising the likelihood of a rewarded trajectory bends the velocity field so that \emph{nearby} seeds flow to the same image; in the limit every seed is mapped to one of the group's best images, which is best-of-$G$ selection compiled into the weights. If this were the only route, two things would follow for an all-or-nothing criterion: \passone on $c$ could approach, but not exceed, the base model's \passk at $k$ of the order of the samples $c$ receives, and a prompt with success probability $p$ would yield a success-free group with probability $(1-p)^G$, from which nothing about the criterion is learned. This is a first-order statement: later groups come from the updated policy. We call such a gain \emph{sharpening}: it concentrates coverage and does not create it.

\paragraph{Across prompts: shared parameters.} The policy is one function $\mu_\theta$ for all prompts. To first order, a gradient step on a group for prompt $c$ changes the log-likelihood of a trajectory $\tau'$ for another prompt $c'$ by $\eta\,\nabla_\theta\log\pi_\theta(\tau'\mid c')^{\top}\nabla_\theta J(c)$. If the base model's failures on $c$ and $c'$ have a common cause, say it lays out objects with little regard to the relation word, then the direction that separates successes from failures on $c$ also separates them on $c'$, the two gradients are aligned, and successes on $c$ raise the success probability on $c'$ even if $c'$ had none. We call this \emph{unlocking}. If instead the failure reflects how the reference model reads one particular word (a purple bear drawn as a toy), nothing learned on other prompts bears on it, and changing the reading is charged by the KL term wherever the word occurs. On this account, unlocking requires (i) a deficit shared by many prompts, (ii) successes on at least some of them, to supply the signal, and (iii) enough of them in the training mix for the shared direction to dominate the update. GenEval's spatial relations satisfy all three: given both objects, the base model arranges them as asked in 39\% of samples (21\% for left and right), 16\% of its samples still succeed, and relations are 41\% of Flow-GRPO's sampled training prompts and 62\% of its distinct ones. Both routes pass through the shared parameters (in our runs a training prompt is visited about once), so a gain on a held-out prompt does not tell them apart; the signatures separate what reached it: more seeds landing where some already did, or images its seeds did not produce.

\paragraph{Signatures and labels.} Sharpening and unlocking leave different traces in quantities that need only inference (\cref{tab:signatures}); $k^\ast$ is the smallest $k$ at which the base model's \passk reaches the post-trained model's \passone. We label a skill operationally: \emph{sharpened} when $k^\ast\le n$ exists and the ceiling rises by at most 0.02, \emph{unlocked} when no $k\le n$ matches, and \emph{mixed} otherwise. For single prompts, and for training in progress, unlocking means zero-coverage prompts gaining coverage. Only the fourth row needs no headroom: where the base model already covers every prompt, sharpening is the only label available.

\begin{table}[t]\centering\small
\resizebox{\linewidth}{!}{\begin{tabular}{lcc}
\toprule
 & Sharpening & Unlocking\\
\midrule
\passone vs.\ base \passk & meets it at small $k^\ast$ & no $k^\ast\le n$\\
ceiling (pass@32) & unchanged & rises\\
zero-coverage prompts & stay unsolved & solved at high rates\\
rate after RL & tracks base rate & ignores base rate\\
best-of-$k^\ast$ with the judge & reproduces gain & cannot\\
\bottomrule
\end{tabular}}
\caption{\textbf{Signatures of sharpening and unlocking}, all measurable with inference alone.}\label{tab:signatures}
\end{table}

\paragraph{Predictions.} The account makes three predictions and leaves one fork (item~2) to the data.
\begin{enumerate}
\item Failures that are a reading held by the KL term stay in place under the penalty and move without it; skills whose failures share a cause and are well represented in training show unlocking (\cref{sec:analysis}, with public checkpoints).
\item If unlocking is a single switch for a relation it should hold at once for objects never seen in training; if the shared direction is assembled from object-level corrections it should appear first on the training vocabulary. Sharpening, which only re-weights what a prompt already covers, should transfer either way (\cref{sec:method}).
\item Sharpening needs only the successes already in a group, unlocking needs the shared direction to accumulate across many prompts; at small budgets a gain should therefore create no coverage on any skill: a flat ceiling and zero-coverage prompts left as they were (\cref{sec:method}).
\item Folding seeds onto the successful images is the same event as losing the images the other seeds produced, so within-prompt diversity should fall wherever \passone rises, the sample-side view of the collapse reported for GRPO-trained generators~\cite{diversegrpo,drift} (\cref{sec:analysis}).
\end{enumerate}

\section{Anatomy of a GenEval Gain}
\label{sec:analysis}

\paragraph{Setup.} We study \sdm~\cite{sd3} and the public LoRA post-trained on GenEval with Flow-GRPO ($G{=}24$, rank 32, the released checkpoint)~\cite{flowgrpo}. Both models are sampled with the same 25-step Euler schedule, guidance 4.5, at $512\times512$, with the deterministic ODE sampler, so the only source of variation is the seed, shared across models. We use the 553 official GenEval prompts~\cite{geneval}, none of which is among Flow-GRPO's 33,199 distinct training prompts, and $n{=}32$ samples per prompt, so \passk is unbiased up to $k{=}32$. Correctness is the strict rule of \cref{sec:background}. It is harsher than the official GenEval rule, which accepts \emph{at least} the requested number of each object: on the same images the official rule gives 0.64 for the base model and 0.94 for the checkpoint, matching the published 0.63 and 0.95~\cite{flowgrpo}, and the strict rule gives 0.51 and 0.91. We report the strict rule, the event the reward is maximized on; under the official rule five of six labels hold (relations: \passone 0.23 to 0.96, pass@32 0.76 to 1.00, no $k^\ast\le32$, all 24 zero-coverage prompts solved at 0.96; other skills: $k^\ast$ of 5 to 14; \cref{sec:rules}). \cref{tab:anatomy} reports per skill the \passk of each model, $k^\ast$, and how many zero-coverage prompts it rescues and at what rate; in total we score more than 180k images.

\begin{table*}[t]\centering\footnotesize\setlength{\tabcolsep}{5pt}
\begin{tabular}{lr cc ccccc ccccc}
\toprule
\multicolumn{14}{l}{\textbf{(a) GenEval, 553 prompts}}\\[1pt]
 & & \multicolumn{2}{c}{\sdm (base)} & \multicolumn{5}{c}{Flow-GRPO} & \multicolumn{5}{c}{Flow-GRPO without KL}\\
\cmidrule(lr){3-4}\cmidrule(lr){5-9}\cmidrule(lr){10-14}
Skill & $N$ & p@1 & p@32 & p@1 & p@32 & $k^\ast$ & resc. & rate$_0$ & p@1 & p@32 & $k^\ast$ & resc. & rate$_0$\\
\midrule
Single object & 80 & 0.84{\scriptsize$\pm$.04} & 1.00 & 0.98{\scriptsize$\pm$.01} & 1.00 & 4 & 0/0 & -- & 0.99{\scriptsize$\pm$.01} & 1.00 & 6 & 0/0 & --\\
Two objects & 99 & 0.50{\scriptsize$\pm$.05} & 0.99 & 0.95{\scriptsize$\pm$.02} & 0.99 & 10 & 1/1 & 0.97 & 0.93{\scriptsize$\pm$.03} & 0.99 & 9 & 1/1 & 0.97\\
Counting & 80 & 0.55{\scriptsize$\pm$.06} & 0.99 & 0.94{\scriptsize$\pm$.03} & 1.00 & 8 & 1/1 & 0.91 & 0.96{\scriptsize$\pm$.02} & 1.00 & 11 & 1/1 & 0.88\\
Colors & 94 & 0.71{\scriptsize$\pm$.06} & 0.96 & 0.89{\scriptsize$\pm$.05} & 0.97 & 5 & 1/4 & 0.16 & 0.93{\scriptsize$\pm$.04} & 0.99 & 14 & 4/4 & 0.73\\
Position & 100 & 0.16{\scriptsize$\pm$.04} & 0.69 & 0.94{\scriptsize$\pm$.02} & 1.00 & $>$32 & 31/31 & 0.93 & 0.92{\scriptsize$\pm$.04} & 0.98 & $>$32 & 31/31 & 0.92\\
Attribute binding & 100 & 0.37{\scriptsize$\pm$.05} & 0.90 & 0.79{\scriptsize$\pm$.05} & 0.98 & 8 & 9/10 & 0.24 & 0.80{\scriptsize$\pm$.05} & 0.98 & 8 & 8/10 & 0.40\\
\midrule
\textbf{All} & 553 & 0.51{\scriptsize$\pm$.03} & 0.92 & 0.91{\scriptsize$\pm$.01} & 0.99 & 31 & 43/47 & 0.72 & 0.92{\scriptsize$\pm$.02} & 0.99 & $>$32 & 45/47 & 0.80\\
\multicolumn{4}{l}{\quad share of the gain inside base coverage} & \multicolumn{5}{c}{85\%} & \multicolumn{5}{c}{84\%}\\
\midrule\midrule
\multicolumn{14}{l}{\textbf{(b) 20-class set, 256 prompts}}\\[1pt]
 & & \multicolumn{2}{c}{\sdm (base)} & \multicolumn{5}{c}{Flow-GRPO (trained on these classes)} & \multicolumn{5}{c}{Run A, ours (classes held out; 14k images)}\\
\cmidrule(lr){3-4}\cmidrule(lr){5-9}\cmidrule(lr){10-14}
Skill & $N$ & p@1 & p@32 & p@1 & p@32 & $k^\ast$ & resc. & rate$_0$ & p@1 & p@32 & $k^\ast$ & resc. & rate$_0$\\
\midrule
Single object & 20 & 0.85{\scriptsize$\pm$.06} & 1.00 & 0.99{\scriptsize$\pm$.01} & 1.00 & 4 & 0/0 & -- & 0.87{\scriptsize$\pm$.05} & 1.00 & 2 & 0/0 & --\\
Two objects & 48 & 0.40{\scriptsize$\pm$.07} & 1.00 & 0.93{\scriptsize$\pm$.04} & 1.00 & 11 & 0/0 & -- & 0.45{\scriptsize$\pm$.07} & 0.96 & 2 & 0/0 & --\\
Counting & 47 & 0.55{\scriptsize$\pm$.08} & 1.00 & 0.95{\scriptsize$\pm$.04} & 1.00 & 12 & 0/0 & -- & 0.65{\scriptsize$\pm$.08} & 1.00 & 2 & 0/0 & --\\
Colors & 47 & 0.72{\scriptsize$\pm$.06} & 1.00 & 0.95{\scriptsize$\pm$.02} & 1.00 & 4 & 0/0 & -- & 0.79{\scriptsize$\pm$.05} & 1.00 & 2 & 0/0 & --\\
Position & 47 & 0.18{\scriptsize$\pm$.06} & 0.68 & 0.89{\scriptsize$\pm$.06} & 1.00 & $>$32 & 15/15 & 0.75 & 0.18{\scriptsize$\pm$.06} & 0.70 & 2 & 4/15 & 0.01\\
Attribute binding & 47 & 0.30{\scriptsize$\pm$.06} & 0.94 & 0.85{\scriptsize$\pm$.05} & 1.00 & 12 & 3/3 & 0.68 & 0.42{\scriptsize$\pm$.06} & 0.98 & 2 & 3/3 & 0.06\\
\midrule
\textbf{All} & 256 & 0.46{\scriptsize$\pm$.04} & 0.93 & 0.92{\scriptsize$\pm$.02} & 1.00 & 27 & 18/18 & 0.74 & 0.53{\scriptsize$\pm$.04} & 0.93 & 2 & 7/18 & 0.02\\
\multicolumn{4}{l}{\quad share of the gain inside base coverage} & \multicolumn{5}{c}{89\%} & \multicolumn{5}{c}{98\%}\\
\bottomrule
\end{tabular}

\caption{\textbf{Anatomy of RL gains per skill} ($n{=}32$ shared seeds per prompt, 25 steps, strict rule; $\pm$ is the half-width of a 95\% bootstrap interval over prompts). $k^\ast$ is the smallest $k$ at which the base model's \passk reaches the model's \passone; resc.\ counts, among the zero-coverage prompts, those the model solves at least once, and rate$_0$ is its mean success rate on all of them. By the rule of \cref{sec:mechanism}, Flow-GRPO sharpens four skills, unlocks spatial relations and is mixed on attribute binding.}\label{tab:anatomy}
\end{table*}

\paragraph{Four skills sharpen.} For single objects, two objects, counting and colours the post-trained model's \passone (0.98, 0.95, 0.94, 0.89) is what the base model reaches with 4, 10, 8 and 5 samples, and pass@32 moves by at most 0.013. By the rule of \cref{sec:mechanism} these skills are sharpened, with a caveat: the base model already covers 347 of their 353 prompts, so the anatomy measures the price of the gain here, not a limit on what RL can do. Counting shows that sharpening does not need many distinct prompts: its 29\% share of the training samples revisits only 160 prompts, yet \passone on held-out counting prompts rises from 0.55 to 0.94, the base model's pass@8. The six uncovered prompts are the only room these skills leave, and they split: three are unlocked outright (``a carrot and a couch'' goes from 0 to 31 of 32) and three colour prompts stay at zero. The latter are not rare events waiting in the noise: on every seed the model commits to a reading the judge rejects. ``A purple bear'' is drawn as a purple toy, which the detector files under \emph{teddy bear} (re-scored against that class, all sampled images pass; \cref{fig:qual}); ``a brown orange'' is drawn orange.

\begin{figure}[t]\centering\includegraphics[width=.9\linewidth]{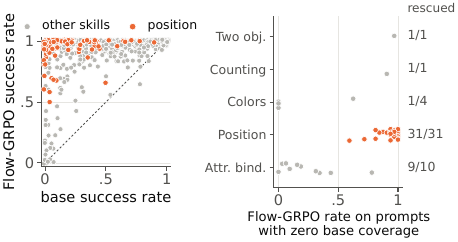}
\caption{\textbf{Where the gain comes from.} Left: per-prompt success rate before and after RL, one dot per prompt; relation prompts in orange. Almost every prompt moves up; the relation prompts end at the top wherever they started, which is coverage created, not concentrated. Right: the post-trained model's success rate on the 47 prompts the base model never solved in 32 samples, by skill, with the number rescued.}\label{fig:buckets}\end{figure}

\paragraph{Spatial relations unlock.} Position is different in every column. The base model succeeds on 16\% of samples and, given 32 attempts, on 69\% of prompts; the post-trained model succeeds on 94\% of samples and 100\% of prompts. No $k\le32$ brings the base model to the post-trained model's \passone. All 31 zero-coverage relation prompts are now solved at a mean rate of 0.93, and they are unremarkable (``a cup left of an umbrella''). Two measurements argue against a rare success made frequent. First, the graded reward splits a relation into drawing both objects and arranging them. On these 31 prompts the base model draws both objects in 28\% of the 128 samples drawn below and arranges them as asked in 5\% of those (56 of 1{,}114), against 48\% (2{,}005 of 4{,}141) on the 69 relation prompts it solves; the post-trained model arranges them as asked in 98\% (923 of 939). The deficit is a layout the model prefers, not an object it cannot draw. Second, the outcome does not depend on what the base model had. Relation prompts it solved 0, 1--2, 3--8 and 9--16 times in 32 end at 0.93, 0.91, 0.97 and 0.95 (31, 21, 23 and 18 prompts); the same groups on the other five skills end at 0.31, 0.44, 0.82 and 0.93 (16, 15, 56 and 91 prompts; \cref{fig:buckets}, left). Re-weighting leaves a prompt's rate tied to where it started; the update that reaches relation prompts ignores it. This criterion needs no headroom, and it is the evidence we weigh most.

\paragraph{Attribute binding is mixed.} Two coloured objects is the hardest skill after RL. The post-trained model's \passone (0.79) is the base model's pass@8, as in sharpening, but pass@32 moves by 0.08 and nine of the ten zero-coverage prompts gain coverage, the ten reaching a mean rate of 0.24. Under \cref{sec:mechanism} this is what a partly shared deficit should look like: binding a colour to the right object is something many prompts need, and the one prompt that stays at zero, ``a black kite and a green bear'', is drawn with a green bear and the kite as a bird of prey.

\paragraph{The gain decomposed.} Flow-GRPO raises the strict score from 0.506 to 0.913. Three accountings of this gain answer three different questions, and we report all of them instead of a single share. \emph{Where does it fall?} 15\% (0.061) on the 47 zero-coverage prompts, 31 of them relations, and 85\% (0.346) on prompts the base model solved at least once; this is a floor for unlocking, because it credits to sharpening every relation prompt with one lucky seed. \emph{Which skill carries it?} Relations 35\%, the other five skills 65\%. \emph{How much does selection buy back?} If each prompt is given the smaller of the base model's \passk and the post-trained model's rate, selecting with the judge recovers 59\%, 73\% and 85\% of the gain at $k{=}4$, 8 and 32; at $k{=}8$ it recovers 90\% on the five skills and 41\% on relations. Forgetting is negligible: one prompt in 553 loses more than 0.1, and none of the 24 the base model always solves drops below 0.96.

\begin{figure*}[t]\centering\includegraphics[width=.88\textwidth]{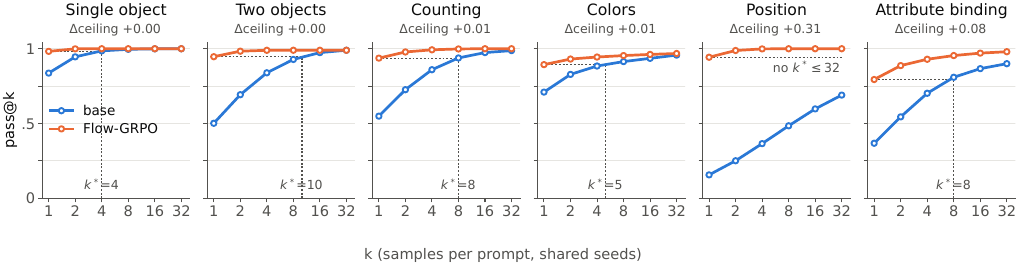}
\caption{\textbf{Two shapes.} \passk over seeds for the base model and the Flow-GRPO checkpoint, per GenEval skill (553 prompts, $n{=}32$, shared seeds). Dotted lines mark $k^\ast$. Four skills: the curves meet by $k{=}32$ under a flat ceiling (sharpening). Relations: no $k\le32$ catches up and the ceiling rises by 0.31 (unlocking).}\label{fig:passk}\end{figure*}

\paragraph{What the headline hides.} Over the whole benchmark the base model needs 31 samples to match one from the post-trained model (\cref{tab:anatomy}a, row \emph{All}; \cref{fig:passk}). No skill needs 31: five need 4 to 10, and for relations 128 do not suffice (below). The aggregate describes none of them. Relations are also where the training mix concentrates: 41\% of the sampled training prompts and 62\% of the distinct ones, against 29\% and 0.5\% for counting and 18\% and 27\% for attribute binding. Counting, with 29\% of the samples, shows no unlocking, but it had no room to show any: the structure of a deficit, its data share and the headroom the base model leaves are confounded in the public checkpoints.

\paragraph{Deeper sampling.} A prompt unsolved in 32 samples may still have a success rate of several percent, so we drew 128 samples from the base model for every relation prompt and for the 16 other zero-coverage prompts. Nineteen of the 47 remain unsolved (eleven relation, five binding and three colour prompts); the other 28 have success rates between 0.8\% and 5.5\%. The post-trained model solves those eleven relation prompts at a mean rate of 0.92. For relations as a whole the base model's pass@128 is 0.89, still below the post-trained model's \passone of 0.94: on this skill one post-trained sample is worth more than 128 base samples. With 40 steps instead of 25 the base model's \passone changes by $+0.02$ and the post-trained model's by less than $0.01$ on a 200-prompt subset, and at $1024\times1024$ the base model's by less than $0.02$ (\cref{sec:budget}).

\paragraph{One judge.} Every success in this paper is the verdict of a detector, which errs in both directions (\cref{fig:qual}, middle). It is also the detector behind the reward Flow-GRPO trained on: selecting with it at inference and training on it are two ways of spending one signal, and our comparison holds that signal fixed. ``Best-of-$k$ recovers the gain'' is thus a statement under this judge, at the price of $k$ samples and $k$ detector calls.

\paragraph{Removing the KL penalty.} The released checkpoint trained without the KL term~\cite{flowgrpo} keeps five of the six labels (\cref{tab:anatomy}a); colours turn mixed (pass@32 $+0.03$), for an instructive reason. Prompts on which the regularized model keeps the base model's reading now satisfy the judge: ``a purple bear'' at 1.00, drawn as an animal instead of a toy, and ``a brown orange'' at 0.97, drawn as a dried slice. But every image takes on the same pale background and the seeds collapse onto near-identical compositions (\cref{fig:qual}), and ten prompts regress by more than 0.1, against one for the regularized model, among them ``a white sandwich'' (0.28 to 0.03; 0.84 with KL). These failures were not unreachable; they were protected, as prediction~1 has it: removing the KL term buys the judge's reading together with a drift the reward does not see.

\paragraph{A second task: text rendering.} We apply the anatomy to the released Flow-GRPO checkpoint trained with an OCR reward~\cite{flowgrpo}, on 200 held-out prompts that ask for a quoted string, scored correct when the string is read back verbatim (\cref{tab:ocr}). Here sharpening and unlocking meet in one skill, mixed by our rule. \passone rises from 0.19 to 0.64 with $k^\ast{=}12$, and 84\% of the gain falls within base coverage. But the ceiling also rises, from 0.79 to 0.97, and 37 of the 42 strings the base model never renders in 32 samples are rescued, the 42 reaching a mean rate of 0.35. The unlocked part grows with the length of the string, from a ceiling change of 0.11 for short strings to 0.26 for long ones: unlocking shows where the base model leaves room for it.

\begin{table}[t]\centering\small\setlength{\tabcolsep}{3pt}
\resizebox{\linewidth}{!}{\begin{tabular}{lrcccccccc}
\toprule
Skill & $N$ & \multicolumn{2}{c}{base} & \multicolumn{2}{c}{RL} & $k^\ast$ & $\Delta$ceiling & zero-cov.\ (rescued)\\
\cmidrule(lr){3-4}\cmidrule(lr){5-6}
 & & p@1 & p@32 & p@1 & p@32 & & p@32 & \\
\midrule
Short string ($\le$13 chars) & 80 & 0.28 & 0.85 & 0.73 & 0.96 & 10 & +0.11 & 12 (9) \\
Medium (14--16) & 58 & 0.16 & 0.79 & 0.65 & 1.00 & 14 & +0.21 & 12 (12) \\
Long ($\ge$17) & 62 & 0.10 & 0.71 & 0.51 & 0.97 & 11 & +0.26 & 18 (16) \\
\midrule
\textbf{All} & 200 & 0.19 & 0.79 & 0.64 & 0.97 & 12 & +0.18 & 42 (37) \\
\bottomrule
\end{tabular}
}
\caption{\textbf{Text rendering} (Flow-GRPO-Text against the base model, 200 OCR prompts, $n{=}32$), by length of the target string. $\Delta$ceiling is the change in pass@32; the last column counts strings the base model never renders and, in parentheses, those rescued. Both signatures are present, and the ceiling rises most where base coverage is lowest.}\label{tab:ocr}
\end{table}

\paragraph{A second prompt set.} On 256 prompts from 20 other object classes, 132 of them in the released checkpoint's training set, the labels are the same (\cref{tab:anatomy}b), though zero-coverage relation prompts end lower than the rest (0.75 against 0.96; \cref{sec:data}); \cref{sec:method} holds these classes out.

\paragraph{Diversity.} Within-prompt diversity, the mean pairwise DINOv2~\cite{dinov2} distance among eight shared-seed samples, falls by a third after RL (0.36 to 0.24 on 200 prompts) and falls on every skill, as prediction~4 expects. Part of the fall is correctness itself, since failures are varied and the measure includes them. It is largest on relations (0.39 to 0.22), where the unlocked skill arrives as one stereotyped layout (\cref{fig:qual}, top).

\begin{figure}[t]\centering\includegraphics[width=.9\linewidth]{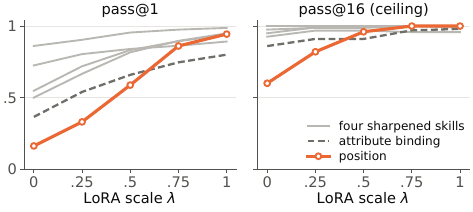}
\caption{\textbf{The LoRA dial.} \passone and the pass@16 ceiling per skill as the Flow-GRPO adapter is scaled by $\lambda$ ($n{=}16$ seeds). Sharpened skills gain \passone under a nearly flat ceiling. For spatial relations the ceiling rises first and \passone follows.}\label{fig:dial}\end{figure}

\paragraph{The LoRA dial.} Scaling the adapter by $\lambda\in\{0,0.25,0.5,0.75,1\}$ interpolates between the two models at no training cost (\cref{fig:dial}). For the four sharpened skills \passone climbs with $\lambda$ while pass@16 moves by at most 0.05. For relations both move, and the ceiling moves first: pass@16 goes 0.60, 0.82, 0.96, 1.00, 1.00 while \passone goes 0.16, 0.33, 0.59, 0.86, 0.94. Half the adapter already gives coverage on nearly every relation prompt. That the ceiling leads is what \passk arithmetic gives any rising rate; what the dial adds is that the whole transition lies along one direction in weight space and can be inspected without any training.


\section{Controlled Training: What Happens First}
\label{sec:method}

The public checkpoints are end points. To see the order in which sharpening and unlocking appear, and what a gain is tied to, we train our own adapters under a small budget and read the same anatomy along the way.

\paragraph{Protocol.} All runs use \sdm with the Flow-GRPO pipeline and its graded GenEval reward on a single GPU, one training seed each (LoRA rank 32, group $G{=}8$, 64 images and on average 1.9 optimizer steps per epoch, KL $\beta{=}0.04$, 10 collection steps; details in \cref{sec:controlled}). Training prompts are the Flow-GRPO GenEval templates restricted to 60 of the 80 object classes~\cite{flowgrpo}; the other 20 classes appear in no training prompt. Run~A trains on all skills (415 updates, 14k images). Run~B trains on relation prompts only and is evaluated after 768, 1{,}300, 1{,}903, 2{,}291 and 2{,}730 updates (26k, 44k, 64k, 78k and 95k images). For scale, the released configuration uses $G{=}24$ and 1{,}152 images per epoch, and Run~B's roughly 12k groups visit each of its 10{,}777 distinct training prompts about once. We evaluate relation prompts with $n{=}32$ seeds shared with the base model, grouped by how many of their two objects are from held-out classes: none (59 prompts), one (39) or both (49).

\paragraph{Small budgets sharpen.} Run~A raises \passone on the 322 official prompts it never saw from 0.51 to 0.56 and on the 256 unseen-object prompts from 0.46 to 0.53. On every skill $k^\ast{=}2$ (\cref{tab:anatomy}b), as it must be for a gain this small. 97--98\% of the gain accrues on prompts the base model already solved, the pooled ceiling moves by 0.01 and 0.00 (single skills by at most 0.06, in both directions), and the zero-coverage prompts reach a mean rate of 0.02, less than one sample in 32~(\cref{tab:method}): what prediction~3 expects, up to sampling noise.

\begin{figure}[t]\centering\includegraphics[width=.9\linewidth]{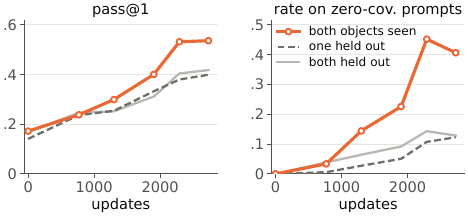}
\caption{\textbf{Sharpening first, unlocking later and object by object.} Relation-only training (Run~B). Left: \passone rises for all three groups of relation prompts. Right: the success rate on prompts with zero base coverage rises fastest when both objects were in the training vocabulary.}\label{fig:traj}\end{figure}

\paragraph{Then coverage appears, on the objects that were practised.} Run~B spends every sample on relations (\cref{fig:traj}). After 26k images its gain is still about one extra draw ($k^\ast$ of 2 or 3) for all three groups, and zero-coverage prompts barely move (rates of 0.03 and 0.04). From there the groups part ways. On prompts whose two objects were both in the training vocabulary, \passone reaches 0.53 from 0.17 after 78k images, $k^\ast$ grows to 9, and 11 of the 13 zero-coverage prompts are solved at least three times in 32 samples, at a mean rate of 0.45; under the base model none of them exceeds 0.055 even at $n{=}128$. On prompts with one or two held-out objects \passone also rises, to 0.38 and 0.40, and their zero-coverage prompts follow at a distance: mean rates of 0.11 and 0.14, with 8 of 16 and 10 of 17 solved three times, up from 2 and 4 one segment earlier. A further 17k images move neither side: at 95k \passone stands at 0.54, 0.40 and 0.42 for the three groups, and the zero-coverage rates at 0.41 against 0.13. The gap to the practised objects is not an accident of which prompts fell in which group (permutation test over prompts, $p{=}0.01$ at 64k, $p<10^{-3}$ at 78k and at 95k; the test does not cover variance across training runs), and class difficulty explains little of it: the released checkpoint, which trained on these classes, takes the 33 prompts to a mean rate of 0.84 and the 13 others to 0.94. Nine of the 59 seen-object prompts never occurred in training and fifty did; the nine reach a \passone of 0.53 at 78k, as do the fifty. What matters is the object vocabulary, not the prompt string: having trained on a prompt adds nothing measurable.

\paragraph{Reading.} Two things follow for these runs. First, the order: sharpening comes first and everywhere, including on objects never seen in training, while unlocking shows only later. An early rise of the reward is thus no evidence that a skill is being acquired. Second, the grain: at this budget unlocking is not a switch thrown for the relation word. It arrives first for the objects the adapter has practised placing and reaches the others more slowly, the second branch of item~2: the shared direction of \cref{sec:mechanism} looks \mbox{assembled} from object-level corrections.

\begin{table}[t]\centering\small
\setlength{\tabcolsep}{2.6pt}
\resizebox{\linewidth}{!}{\begin{tabular}{lcccc|cccc}
\toprule
 & \multicolumn{4}{c|}{both objects seen ($N{=}59$)} & \multicolumn{4}{c}{both held out ($N{=}49$)}\\
 & p@1 & $k^\ast$ & resc. & rate$_0$ & p@1 & $k^\ast$ & resc. & rate$_0$\\
\midrule
\sdm base & 0.17 & -- & 0/13 & 0 & 0.17 & -- & 0/17 & 0\\
Run A, all skills (14k) & 0.19 & 2 & 1/13 & 0.02 & 0.17 & 2 & 0/17 & 0.01\\
Run B, relations (26k) & 0.24 & 2 & 2/13 & 0.03 & 0.25 & 2 & 4/17 & 0.04\\
Run B, relations (44k) & 0.30 & 3 & 9/13 & 0.14 & 0.25 & 2 & 5/17 & 0.06\\
Run B, relations (64k) & 0.40 & 5 & 10/13 & 0.23 & 0.31 & 3 & 4/17 & 0.09\\
Run B, relations (78k) & 0.53 & 9 & 11/13 & 0.45 & 0.40 & 5 & 10/17 & 0.14\\
Run B, relations (95k) & 0.54 & 9 & 10/13 & 0.41 & 0.42 & 5 & 8/17 & 0.13\\
\midrule
Flow-GRPO, released & 0.94 & $>$32 & 13/13 & 0.94 & 0.89 & $>$32 & 16/17 & 0.78\\
\bottomrule
\end{tabular}
}
\caption{\textbf{Controlled runs on relation prompts}, by whether the two objects were in the 60-class training vocabulary ($n{=}32$; image budget in parentheses). ``resc.'' here counts zero-coverage prompts solved at least three times in 32 samples, a stricter count than in \cref{tab:anatomy} because small rates are the question; rate$_0$ is the mean success rate on them.}\label{tab:method}
\end{table}

\section{Related Work}
\label{sec:related}

\paragraph{RL post-training for diffusion and flow models.} DDPO and DPOK~\cite{ddpo,dpok} introduced policy gradients for diffusion; Flow-GRPO~\cite{flowgrpo} and DanceGRPO~\cite{dancegrpo} brought group-relative optimization to flow models, followed by variants that reshape timestep credit~\cite{tempflow}, confine optimization to a sliding window of timesteps~\cite{mixgrpo}, replace likelihoods with forward-process contrast~\cite{diffusionnft}, or move offline~\cite{gdro}. These works report reward gains; none asks which part of the gain was available from the base model's own samples. We analyse Flow-GRPO only.

\paragraph{Diversity collapse.} Several works observe that GRPO-trained generators homogenize and propose diversity-aware rewards or regularizers~\cite{diversegrpo,drift,gardo}. Sharpening accounts for part of it: folding seeds onto the images a few of them produced is the same event as losing the others.

\paragraph{Sharpening versus discovery in LLMs.} Yue~\etal~\cite{yue2025does} showed with \passk that RLVR narrows LLM reasoning to paths already present in the base model; ProRL~\cite{prorl} reports the opposite with longer training, and several objectives aim to preserve \passk~\cite{nsr,passktraining,unlikeliness}. Our data say both sides describe real behaviour, on different skills of one model after one run: a benchmark aggregates skills with different coverage, and the aggregate takes the dominant shape.

\paragraph{Noise and inference-time search.} ``Golden'' seeds~\cite{goldennoise}, noise generators trained with the model frozen~\cite{noiseppo}, rejection-sampling fine-tuning~\cite{graft} and test-time search~\cite{inferencescaling} are all consistent with sharpening: the base model already covers the target.

\section{Discussion and Conclusion}
\label{sec:conclusion}

A 0.63 that becomes 0.95 is a real improvement in what a user sees at the first attempt, and nothing here diminishes it. What we dispute is that the number measures one thing. On this model it measures two: a concentration of images the base model already drew, of which selection among 8 base samples with the same judge recovers nine tenths outside relations, and coverage of prompts the base model did not solve, which 128 base samples do not buy. If a skill sharpens, the levers are at inference: samples and a verifier. If it can be unlocked, they are the training mix, with many prompts that share the deficit and enough successes among them, and the KL budget. We recommend that reported reward gains come with two numbers per skill, the $k^\ast$ at which the base model's \passk matches the gain and the number of zero-coverage prompts rescued, each measured inside and outside the training vocabulary.

\paragraph{Limitations.} Our evidence is one base model and one algorithm: three released Flow-GRPO checkpoints and our own small runs. The account of \cref{sec:mechanism} is tested only through its predictions: we do not measure gradient alignment, and a method without SDE likelihoods~\cite{diffusionnft} would test its flow-specific step. Success is the verdict of the judge behind the reward, without human evaluation, so selection and training are compared under that judge only. Coverage is budgeted and measured with the ODE sampler: no success in 128 samples bounds a rate below 0.023, not at zero. On four GenEval skills the base model leaves no room to observe unlocking, and deficit structure, data share and headroom are confounded in the public checkpoints. The controlled study is single-GPU and single-seed, with 39 to 59 prompts per group, and stops where unlocking is under way: held-out objects have begun to follow the practised ones by 78k samples, the next 17k move neither, for reasons we do not know, and whether the gap closes at the released checkpoint's budget is open.

\section*{LLM Usage Statement}
An LLM-based agent (Claude Code, by Anthropic) was used in this work, under the authors' direction, in both the research and the writing. \emph{Research.} The authors led the research direction. The research question was developed through discussions between the authors and the agent, which also helped survey related work. The agent wrote the experiment code, ran the experiments, and produced the tables and figures from the raw outputs. \emph{Writing.} The authors specified the structure and key points of each section; the agent drafted and revised the text accordingly, and the authors reviewed and edited the final text. \emph{Checking.} The authors and a separate agent session checked the draft against the raw results and verified every bibliography entry against its published source. Issues identified by the authors were corrected with the agent's assistance. The authors are responsible for all content of this paper.  %
{
    \small
    \bibliographystyle{ieeenat_fullname}
    \bibliography{main}
}
\clearpage
\appendix
\twocolumn[\centering\Large\textbf{Appendix}\vspace{1.5em}]
\setcounter{table}{0}\setcounter{figure}{0}
\renewcommand{\thetable}{S\arabic{table}}\renewcommand{\theHtable}{S\arabic{table}}
\renewcommand{\thefigure}{S\arabic{figure}}\renewcommand{\theHfigure}{S\arabic{figure}}
\setcounter{topnumber}{4}\setcounter{bottomnumber}{3}\setcounter{totalnumber}{7}\setcounter{dbltopnumber}{3}
\renewcommand{\topfraction}{0.95}\renewcommand{\bottomfraction}{0.85}\renewcommand{\textfraction}{0.04}
\renewcommand{\floatpagefraction}{0.85}\renewcommand{\dbltopfraction}{0.95}\renewcommand{\dblfloatpagefraction}{0.85}
\raggedbottom
This appendix gives the details the main text points to (the training configuration, the 40-step and 1024-pixel checks) and finer cuts of the experiments it reports. It contains no new experiment: every table is computed from the samples behind the tables of the main text, and every image comes from the runs evaluated there. Tables and figures of the appendix carry the prefix S.

\section{The judge, the reward and the published scores}
\label{sec:rules}

\paragraph{Two rules on one detector.} GenEval scores an image by running an object detector and checking the clauses of the prompt~\cite{geneval}. Its official rule accepts an image when the detector finds \emph{at least} the requested number of each object and the colour and relation checks pass on the most confident detections. The reward released with Flow-GRPO~\cite{flowgrpo} scores the same detections clause by clause and reaches its maximum only when each object is found \emph{exactly} the requested number of times; this is the strict rule of the main text. Every image the strict rule accepts is accepted by the official rule, without exception among the 53k images of the three GenEval runs, so the two rules differ by the images that contain surplus objects: 13\% of the base model's images and under 3\% of each checkpoint's.

\Cref{tab:rules} scores the same images under both rules. Under the official rule the base model scores 0.64 and the Flow-GRPO checkpoint 0.94 averaged over prompts, as quoted in \cref{sec:analysis}, and 0.65 and 0.94 with GenEval's average over skills, against the published 0.63 and 0.95. The distance between our headline numbers (0.51 and 0.91) and the published ones is therefore the rule; \cref{sec:budget} shows that it is neither the number of steps nor the resolution. The rule matters where surplus objects are common, most of all on two objects (0.50 strict against 0.78 official for the base model); counting is unaffected because its official check already bounds the count from above.

\begin{table}[H]\centering\small\setlength{\tabcolsep}{3pt}
\resizebox{\linewidth}{!}{\begin{tabular}{lr cc cc cc}
\toprule
 & & \multicolumn{2}{c}{\sdm (base)} & \multicolumn{2}{c}{Flow-GRPO} & \multicolumn{2}{c}{Flow-GRPO without KL}\\
\cmidrule(lr){3-4}\cmidrule(lr){5-6}\cmidrule(lr){7-8}
Skill & $N$ & strict & official & strict & official & strict & official\\
\midrule
Single object & 80 & 0.84 & 0.98 & 0.98 & 1.00 & 0.99 & 1.00\\
Two objects & 99 & 0.50 & 0.78 & 0.95 & 0.99 & 0.93 & 0.98\\
Counting & 80 & 0.55 & 0.55 & 0.94 & 0.94 & 0.96 & 0.96\\
Colours & 94 & 0.71 & 0.80 & 0.89 & 0.91 & 0.93 & 0.94\\
Position & 100 & 0.16 & 0.23 & 0.94 & 0.96 & 0.92 & 0.95\\
Attribute binding & 100 & 0.37 & 0.54 & 0.79 & 0.85 & 0.80 & 0.85\\
\midrule
Mean over prompts & 553 & 0.51 & 0.64 & 0.91 & 0.94 & 0.92 & 0.94\\
Mean over skills (GenEval score) &  & 0.52 & 0.65 & 0.92 & 0.94 & 0.92 & 0.95\\
Published~\cite{flowgrpo} & & & 0.63 & & 0.95 & & --\\
\bottomrule
\end{tabular}}
\caption{\textbf{One set of images, two rules} (\passone, 553 prompts, $n{=}32$). The strict rule requires exactly the requested number of each object, the official rule at least that number. The published scores are reproduced under the official rule.}\label{tab:rules}
\end{table}

\paragraph{The graded reward.} The released reward is the mean of per-clause credits. A requested object contributes $1-|c-\hat c|/c$ for requested count $c$ and detected count $\hat c$, which is negative when the detector finds several surplus instances. A colour clause contributes the fraction of the requested objects that have the requested colour. A relation clause contributes 1 when it holds; when it fails the released code records a 0 and a 1, so that both objects in another arrangement score $3/4$. For a relation prompt this gives the levels of \cref{tab:reward}. An image is strictly correct exactly when its reward is 1, which we verified on every image we scored.

\begin{table}[H]\centering\small\setlength{\tabcolsep}{4pt}
\resizebox{\linewidth}{!}{\begin{tabular}{llccc}
\toprule
Reward & Image contains & base & Flow-GRPO & no KL\\
\midrule
1 & both objects once, arranged as asked & 15.7\% & 94.3\% & 92.3\%\\
0.75 & both objects once, another arrangement & 24.3\% & 1.6\% & 2.3\%\\
1/3 & one object missing or doubled & 38.3\% & 3.6\% & 5.0\%\\
$\le 0$ & two or more counting errors & 21.7\% & 0.5\% & 0.4\%\\
\bottomrule
\end{tabular}}
\caption{\textbf{Reward levels on relation prompts} and the share of images at each level (100 prompts, $n{=}32$). Only the step from 0.75 to 1 concerns the arrangement, and it is all-or-nothing.}\label{tab:reward}
\end{table}

\begin{table*}[!t]\centering\footnotesize\setlength{\tabcolsep}{5pt}
\begin{tabular}{lr cc ccccc ccccc}
\toprule
 & & \multicolumn{2}{c}{\sdm (base)} & \multicolumn{5}{c}{Flow-GRPO} & \multicolumn{5}{c}{Flow-GRPO without KL}\\
\cmidrule(lr){3-4}\cmidrule(lr){5-9}\cmidrule(lr){10-14}
Skill & $N$ & p@1 & p@32 & p@1 & p@32 & $k^\ast$ & resc. & rate$_0$ & p@1 & p@32 & $k^\ast$ & resc. & rate$_0$\\
\midrule
Single object & 80 & 0.98 & 1.00 & 1.00 & 1.00 & 13 & 0/0 & -- & 1.00 & 1.00 & 15 & 0/0 & --\\
Two objects & 99 & 0.78 & 1.00 & 0.99 & 1.00 & 14 & 0/0 & -- & 0.98 & 1.00 & 9 & 0/0 & --\\
Counting & 80 & 0.55 & 0.99 & 0.94 & 1.00 & 8 & 1/1 & 0.91 & 0.96 & 1.00 & 11 & 1/1 & 0.88\\
Colours & 94 & 0.80 & 0.97 & 0.91 & 0.97 & 5 & 0/3 & 0.00 & 0.94 & 0.99 & 13 & 3/3 & 0.69\\
Position & 100 & 0.23 & 0.76 & 0.96 & 1.00 & $>$32 & 24/24 & 0.96 & 0.95 & 0.98 & $>$32 & 24/24 & 0.91\\
Attribute binding & 100 & 0.54 & 0.96 & 0.85 & 0.98 & 6 & 3/4 & 0.06 & 0.85 & 0.99 & 6 & 3/4 & 0.31\\
\midrule
\textbf{All} & 553 & 0.64 & 0.94 & 0.94 & 0.99 & 30 & 28/32 & 0.75 & 0.94 & 0.99 & $>$32 & 31/32 & 0.81\\
\multicolumn{4}{l}{\quad share of the gain inside base coverage} & \multicolumn{5}{c}{86\%} & \multicolumn{5}{c}{85\%}\\
\bottomrule
\end{tabular}
\caption{\textbf{The anatomy of \cref{tab:anatomy}a under the official GenEval rule} (same images, same columns). Spatial relations remain the only skill that no $k\le32$ matches; for Flow-GRPO, attribute binding, mixed under the strict rule, is sharpened under this one.}\label{tab:official}
\end{table*}

\begin{table*}[!t]\centering\footnotesize\setlength{\tabcolsep}{5pt}
\begin{tabular}{lrr ccc ccc ccc}
\toprule
 & & & \multicolumn{3}{c}{\sdm (base)} & \multicolumn{3}{c}{Flow-GRPO} & \multicolumn{3}{c}{Flow-GRPO without KL}\\
\cmidrule(lr){4-6}\cmidrule(lr){7-9}\cmidrule(lr){10-12}
Prompts & $N$ & zero-cov. & objects & arranged & correct & objects & arranged & correct & objects & arranged & correct\\
\midrule
left of & 19 & 8 & 0.37 & 0.21 & 0.08 & 0.98 & 1.00 & 0.98 & 0.95 & 1.00 & 0.95\\
right of & 27 & 11 & 0.36 & 0.21 & 0.08 & 0.96 & 0.96 & 0.92 & 0.93 & 0.95 & 0.89\\
above & 26 & 5 & 0.38 & 0.62 & 0.23 & 0.96 & 0.99 & 0.95 & 0.96 & 0.96 & 0.93\\
below & 28 & 7 & 0.49 & 0.45 & 0.22 & 0.94 & 0.99 & 0.93 & 0.94 & 1.00 & 0.94\\
\midrule
all relations & 100 & 31 & 0.40 & 0.39 & 0.16 & 0.96 & 0.98 & 0.94 & 0.95 & 0.98 & 0.92\\
zero-coverage & 31 & 31 & 0.25 & 0.00 & 0.00 & 0.95 & 0.98 & 0.93 & 0.94 & 0.98 & 0.92\\
covered & 69 & 0 & 0.47 & 0.49 & 0.23 & 0.96 & 0.98 & 0.95 & 0.95 & 0.97 & 0.92\\
\bottomrule
\end{tabular}
\caption{\textbf{Objects and arrangement on relation prompts} ($n{=}32$). \emph{objects}: share of images with both objects detected exactly once; \emph{arranged}: share of those in which the requested relation holds; \emph{correct}: their product, the strict \passone. Zero-coverage and covered refer to the base model's 32 samples.}\label{tab:placement}
\end{table*}

Two consequences matter for the account of \cref{sec:mechanism}. First, the reward is not constant on prompts the base model never solves: on 45 of the 47 zero-coverage prompts the 32 base samples receive at least two different rewards, so a group drawn for such a prompt does carry a gradient. Second, that gradient points at the lower clauses. On relation prompts 24\% of the base model's images sit at 0.75 and 38\% at $1/3$, and the step from 0.75 to 1, the arrangement itself, is all-or-nothing: a group without a correctly arranged image can teach the model to draw both objects once, and nothing about where to put them.

\paragraph{The anatomy under the official rule.} \Cref{tab:official} repeats the anatomy of \cref{tab:anatomy}a under the official rule. Relations remain the one skill that no $k\le32$ matches: \passone goes from 0.23 to 0.96, the ceiling from 0.76 to 1.00, and all 24 prompts without coverage under this rule are solved, at a mean rate of 0.96. The other skills are matched at $k^\ast$ between 5 and 14. The two largest values belong to single and two objects, where the post-trained model is within a point of 1 under this rule (1.00 and 0.99) and $k^\ast$ measures the last percent. One label changes for Flow-GRPO: attribute binding is mixed under the strict rule (ceiling $+0.08$) and sharpened under the official one ($+0.02$), because six of its ten strict zero-coverage prompts fail only on a surplus object and have coverage under the official rule. The share of the gain that falls on prompts the base model already solves is 86\%, against 85\% under the strict rule.

\section{Relations: objects and arrangement}
\label{sec:placement}

\begin{figure*}[!t]\centering\includegraphics[width=.93\textwidth]{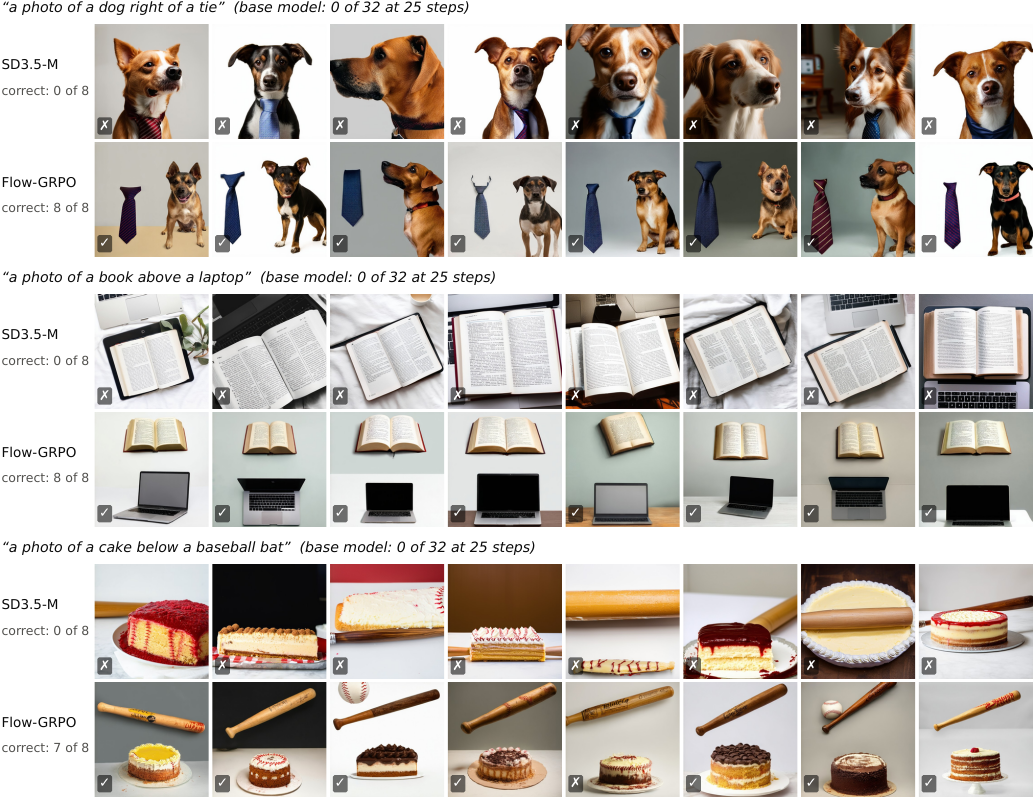}
\caption{\textbf{Three relation prompts the base model never solves in 32 samples} (1, 1 and 2 successes in 128), eight shared seeds, 40 steps. The mark in each image is the judge's verdict. The base model fuses the two objects into one scene or leaves one out (the judge finds both in 5, 2 and 0 of these eight images); the post-trained model separates them, arranges them as asked, and repeats one layout across seeds.}\label{fig:rel}\end{figure*}

The reward levels split the relation skill into two factors with no judge beyond the one already used: drawing both objects exactly once (a reward of at least 0.75) and, given that, arranging them as asked (a reward of 1). \Cref{tab:placement} reports both by relation word. The base model draws both objects in 40\% of its relation samples and arranges them as asked in 39\% of those; the conditional rate is 21\% for \emph{left of} and for \emph{right of}, 62\% for \emph{above} and 45\% for \emph{below}. The post-trained model draws both objects in 96\% of its samples and arranges them as asked in 98\%, with little difference between relation words. The gain on relations is thus the product of two gains: one on the presence of the objects (0.40 to 0.96), which is the two-object skill and has the sharpening signature there (\cref{tab:anatomy}), and one on the arrangement (0.39 to 0.98).

The zero-coverage prompts isolate the second factor. On these 31 prompts the base model draws both objects in 25\% of its 32 samples, 251 images, and arranges none of them as asked. With 128 samples it draws both objects in 28\%, 1{,}114 images, and arranges 56 of them as asked, which is 5\%, against 48\% (2{,}005 of 4{,}141) on the 69 covered prompts. Both factors are lower than on the covered prompts, the first by less than half (0.28 against 0.47) and the second tenfold: when the objects are there, the layout is not. After RL the two groups cannot be told apart: 0.95 and 0.98 on the zero-coverage prompts, 0.96 and 0.98 on the covered ones.

\Cref{fig:rel} shows three of these prompts. For ``a dog right of a tie'' the base model draws a dog wearing a tie on most seeds: the judge finds both objects in 30 of the 32 samples of the main run and the requested arrangement in none. For ``a book above a laptop'' it draws a book lying on a laptop, which the book often hides (both objects are found in 12 of 32). For ``a cake below a baseball bat'' the judge never finds both objects. The three cover what the aggregate contains: two objects fused into one scene, one object hidden, one object missing. The post-trained model draws the two objects apart on a plain background, in the requested arrangement and in nearly the same composition on every seed. This is the stereotyped layout that the diversity measurement of \cref{sec:analysis} picks up (\cref{sec:diversity}).

\section{Sampling budget and sampler}
\label{sec:budget}

\paragraph{Deep sampling.} \Cref{tab:deepcurve} gives the base model's \passk on relations estimated from 128 samples per prompt. It reaches 0.72 at $k{=}32$ (0.69 on the first 32 seeds, as in \cref{tab:anatomy}), 0.81 at 64 and 0.89 at 128, and stays below the post-trained model's \passone of 0.94 at every budget we measured. \Cref{tab:deep} lists the 47 zero-coverage prompts with the base model's successes in 128 samples and each checkpoint's in 32. Nineteen prompts stay unsolved at 128 samples. Eleven are relation prompts, which Flow-GRPO then solves at a mean rate of 0.92; five are binding prompts (0.12) and three are colour prompts (0.00), the readings held in place by the KL term (\cref{sec:kl}). The other 28 prompts have between 1 and 7 successes in 128 samples.

\begin{table}[H]\centering\small\setlength{\tabcolsep}{3.2pt}
\begin{tabular}{l cccccccc}
\toprule
$k$ & 1 & 2 & 4 & 8 & 16 & 32 & 64 & 128\\
\midrule
base \passk & 0.16 & 0.25 & 0.36 & 0.48 & 0.60 & 0.72 & 0.81 & 0.89\\
\bottomrule
\end{tabular}
\caption{\textbf{Base-model \passk on relations from 128 samples per prompt} (100 prompts). The post-trained model's \passone is 0.94.}\label{tab:deepcurve}
\end{table}

\begin{table*}[!t]\centering\scriptsize\setlength{\tabcolsep}{4pt}
\begin{tabular}{llrrr @{\hspace{14pt}} llrrr}
\toprule
Prompt (``a photo of \ldots'') & skill & base & F-G & no KL & Prompt (``a photo of \ldots'') & skill & base & F-G & no KL\\
 & & /128 & /32 & /32 & & & /128 & /32 & /32\\
\midrule
a brown orange & colour & 0 & 0 & 31 & a refrigerator below a scissors & relation & 1 & 30 & 32\\
a purple bear & colour & 0 & 0 & 32 & a train right of a dining table & relation & 1 & 30 & 31\\
a white kite & colour & 0 & 0 & 3 & a wine glass above a kite & relation & 1 & 26 & 30\\
a bed right of a sports ball & relation & 0 & 30 & 32 & a zebra right of a bed & relation & 1 & 30 & 31\\
a carrot left of an orange & relation & 0 & 32 & 29 & a purple suitcase and an orange pizza & binding & 1 & 5 & 2\\
a frisbee right of a motorcycle & relation & 0 & 32 & 32 & a cake below a baseball bat & relation & 2 & 28 & 26\\
a hair drier below an elephant & relation & 0 & 32 & 32 & a truck left of a refrigerator & relation & 2 & 31 & 32\\
a hot dog right of a skateboard & relation & 0 & 30 & 32 & a blue pizza and a yellow baseball glove & binding & 2 & 25 & 20\\
a knife right of a suitcase & relation & 0 & 31 & 32 & a brown hot dog and a purple pizza & binding & 2 & 11 & 27\\
a refrigerator above a baseball bat & relation & 0 & 32 & 28 & an orange handbag and a green carrot & binding & 2 & 2 & 25\\
a tie right of a baseball bat & relation & 0 & 30 & 3 & four benchs & count & 3 & 29 & 28\\
a tie right of a motorcycle & relation & 0 & 30 & 32 & a bed right of a frisbee & relation & 3 & 31 & 32\\
a toilet left of a kite & relation & 0 & 27 & 8 & a tv below a cow & relation & 3 & 26 & 31\\
an elephant below a horse & relation & 0 & 19 & 32 & a red skis and a brown tie & binding & 3 & 14 & 0\\
a black kite and a green bear & binding & 0 & 0 & 0 & a carrot and a couch & two obj. & 4 & 31 & 31\\
a brown oven and a purple train & binding & 0 & 3 & 12 & a cup left of an umbrella & relation & 4 & 32 & 32\\
a pink dining table and a black sandwich & binding & 0 & 10 & 2 & a truck left of a baseball bat & relation & 4 & 31 & 32\\
a red train and a purple bear & binding & 0 & 1 & 13 & a bus below a toothbrush & relation & 5 & 31 & 30\\
a white handbag and a purple bed & binding & 0 & 6 & 27 & a cow right of a laptop & relation & 5 & 32 & 32\\
a book above a laptop & relation & 1 & 31 & 32 & a cell phone left of a tennis racket & relation & 6 & 32 & 32\\
a computer mouse left of a bench & relation & 1 & 31 & 32 & a couch below a cup & relation & 6 & 30 & 31\\
a dog right of a tie & relation & 1 & 29 & 31 & a purple carrot & colour & 7 & 20 & 27\\
a fork above a hair drier & relation & 1 & 32 & 32 & a parking meter above a broccoli & relation & 7 & 23 & 31\\
a hot dog left of a suitcase & relation & 1 & 32 & 32 &  & & & & \\
\bottomrule
\end{tabular}
\caption{\textbf{The 47 zero-coverage prompts}, sorted by the base model's successes in 128 samples. F-G: Flow-GRPO; no KL: Flow-GRPO without KL, both out of 32.}\label{tab:deep}
\end{table*}

\paragraph{Steps and resolution.} \Cref{tab:sampler} compares samplers on 200 of the official prompts, drawn from the 322 that do not occur in our 60-class training split (\cref{sec:data}), which explains the skill mix. Going from 25 to 40 steps moves the base model from 0.50 to 0.52 and leaves the post-trained model at 0.91. Generating at $1024\times1024$ moves the base model to 0.52 (0.501 to 0.516). Under the official rule the same three base-model runs score 0.65, 0.66 and 0.66 and both post-trained runs 0.94. Neither the number of steps nor the resolution accounts for the distance to the published 0.63; the rule does (\cref{tab:rules}). \Cref{tab:steps40} gives the anatomy at 40 steps with the 8 samples available there. Relations go from 0.18 to 0.94 and all 14 relation prompts unsolved in 8 base samples are solved, at a mean rate of 0.94. With $n{=}8$, $k^\ast$ is resolved only up to 8 and zero coverage means no success in 8 samples, so the remaining columns are coarser than those of \cref{tab:anatomy}: counting and colours, with 14 and 10 prompts here, show $k^\ast{>}8$ where \cref{tab:anatomy} has 8 and 5, the base model's pass@8 being 0.86 and 0.90 against a post-trained \passone of 0.89 and 0.91.

\begin{table}[H]\centering\small\setlength{\tabcolsep}{3.5pt}
\resizebox{\linewidth}{!}{\begin{tabular}{lr ccc cc}
\toprule
 & & \multicolumn{3}{c}{\sdm (base)} & \multicolumn{2}{c}{Flow-GRPO}\\
\cmidrule(lr){3-5}\cmidrule(lr){6-7}
Skill & $N$ & 25 steps & 40 steps & 25 steps, $1024^2$ & 25 steps & 40 steps\\
\midrule
Single object & 50 & 0.84 & 0.89 & 0.88 & 0.99 & 0.96\\
Two objects & 34 & 0.42 & 0.45 & 0.49 & 0.94 & 0.94\\
Counting & 14 & 0.54 & 0.49 & 0.61 & 0.90 & 0.89\\
Colours & 10 & 0.72 & 0.72 & 0.65 & 0.91 & 0.91\\
Position & 33 & 0.17 & 0.18 & 0.19 & 0.95 & 0.94\\
Attribute binding & 59 & 0.40 & 0.43 & 0.36 & 0.80 & 0.82\\
\midrule
\textbf{All}, strict rule & 200 & 0.50 & 0.52 & 0.52 & 0.91 & 0.91\\
\textbf{All}, official rule & 200 & 0.65 & 0.66 & 0.66 & 0.94 & 0.94\\
samples per prompt & & 16 & 8 & 4 & 32 & 8\\
\bottomrule
\end{tabular}}
\caption{\textbf{Sampler checks} (\passone on a 200-prompt subset; strict rule unless stated). Steps and resolution move the base model by 0.02 or less.}\label{tab:sampler}
\end{table}

\begin{table}[H]\centering\small\setlength{\tabcolsep}{4pt}
\resizebox{\linewidth}{!}{\begin{tabular}{lr cc cc ccc}
\toprule
 & & \multicolumn{2}{c}{base} & \multicolumn{5}{c}{Flow-GRPO}\\
\cmidrule(lr){3-4}\cmidrule(lr){5-9}
Skill & $N$ & p@1 & p@8 & p@1 & p@8 & $k^\ast$ & resc. & rate$_0$\\
\midrule
Single object & 50 & 0.89 & 0.98 & 0.96 & 1.00 & 3 & 1/1 & 0.88\\
Two objects & 34 & 0.45 & 0.88 & 0.94 & 1.00 & $>$8 & 4/4 & 0.84\\
Counting & 14 & 0.49 & 0.86 & 0.89 & 1.00 & $>$8 & 2/2 & 0.38\\
Colours & 10 & 0.72 & 0.90 & 0.91 & 1.00 & $>$8 & 1/1 & 0.25\\
Position & 33 & 0.18 & 0.58 & 0.94 & 1.00 & $>$8 & 14/14 & 0.94\\
Attribute binding & 59 & 0.43 & 0.85 & 0.82 & 0.98 & 7 & 8/9 & 0.50\\
\midrule
\textbf{All} & 200 & 0.52 & 0.84 & 0.91 & 0.99 & $>$8 & 30/31 & 0.74\\
\bottomrule
\end{tabular}}
\caption{\textbf{Anatomy at 40 steps} (200 prompts, $n{=}8$, strict rule). Zero coverage here means no success in 8 base samples.}\label{tab:steps40}
\end{table}

\section{Accounting for the gain}
\label{sec:gain}

\cref{sec:analysis} reports three accountings of the gain; this section gives each in full, for both GenEval checkpoints.

\paragraph{By skill and by coverage.} \Cref{tab:gainskill} splits the \passone gain by skill and by whether the prompt had base coverage. Relations carry 34.9\% of the Flow-GRPO gain, 12.8 points of it on zero-coverage prompts. The five other skills carry 65.1\%, of which 2.2 points fall on zero-coverage prompts. Without KL the split is nearly the same, with slightly more of the gain beyond coverage on colours and binding, where the released readings are (\cref{sec:kl}).

\paragraph{By selection.} \Cref{tab:bestofk} gives, for each $k$, the share of the gain recovered when each prompt is credited with the smaller of the base model's \passk and the post-trained model's rate. On the five other skills best-of-4 recovers 76\% of the Flow-GRPO gain and best-of-8 recovers 90\%. On relations best-of-32 recovers 63\%.

\begin{table}[H]\centering\small\setlength{\tabcolsep}{4pt}
\resizebox{\linewidth}{!}{\begin{tabular}{l ccc ccc}
\toprule
 & \multicolumn{3}{c}{Flow-GRPO} & \multicolumn{3}{c}{Flow-GRPO without KL}\\
\cmidrule(lr){2-4}\cmidrule(lr){5-7}
Skill & share & within & beyond & share & within & beyond\\
\midrule
Single object & 5.2 & 5.2 & 0.0 & 5.4 & 5.4 & 0.0\\
Two objects & 19.6 & 19.1 & 0.4 & 18.6 & 18.2 & 0.4\\
Counting & 13.8 & 13.4 & 0.4 & 14.3 & 13.9 & 0.4\\
Colours & 7.7 & 7.4 & 0.3 & 9.0 & 7.8 & 1.3\\
Position & 34.9 & 22.1 & 12.8 & 33.5 & 21.0 & 12.5\\
Attribute binding & 18.9 & 17.8 & 1.1 & 19.1 & 17.4 & 1.8\\
\midrule
\textbf{All} & 100.0 & 85.0 & 15.0 & 100.0 & 83.6 & 16.4\\
\bottomrule
\end{tabular}}
\caption{\textbf{Share of the \passone gain by skill} (percent of the total gain, 0.407 for Flow-GRPO and 0.413 without KL), split into prompts within and beyond base coverage.}\label{tab:gainskill}
\end{table}

\begin{table}[H]\centering\small\setlength{\tabcolsep}{4pt}
\resizebox{\linewidth}{!}{\begin{tabular}{ll ccccc}
\toprule
 & & \multicolumn{5}{c}{gain recovered by best-of-$k$}\\
\cmidrule(lr){3-7}
Model & Prompts & $k{=}2$ & $k{=}4$ & $k{=}8$ & $k{=}16$ & $k{=}32$\\
\midrule
Flow-GRPO & all & 34\% & 59\% & 73\% & 81\% & 85\%\\
Flow-GRPO & other five skills & 46\% & 76\% & 90\% & 95\% & 97\%\\
Flow-GRPO & relations & 12\% & 26\% & 41\% & 54\% & 63\%\\
\midrule
no KL & all & 31\% & 55\% & 70\% & 78\% & 84\%\\
no KL & other five skills & 41\% & 70\% & 84\% & 91\% & 94\%\\
no KL & relations & 12\% & 26\% & 41\% & 53\% & 63\%\\
\bottomrule
\end{tabular}}
\caption{\textbf{What selection buys back.} Each prompt is credited with the smaller of the base model's \passk and the post-trained model's success rate.}\label{tab:bestofk}
\end{table}

\paragraph{By base success count.} \Cref{tab:buckets} is the table behind \cref{fig:buckets} (left). For relation prompts the rate after RL does not depend on the number of base successes: it lies between 0.91 and 1.00 in every bin, and at 0.93 for the 31 prompts with none. For the other skills it rises with the base count, from 0.31 to 0.97. The official rule sharpens the contrast (0.14 to 0.99 against 0.95 to 0.99). Without KL the dependence weakens on the other skills, 0.55 instead of 0.31 at zero base successes, which is the release of the readings discussed in \cref{sec:kl}.

\begin{table}[H]\centering\small\setlength{\tabcolsep}{3pt}
\resizebox{\linewidth}{!}{\begin{tabular}{ll ccccc}
\toprule
 & & \multicolumn{5}{c}{base successes in 32 samples}\\
\cmidrule(lr){3-7}
Model & Prompts & 0 & 1--2 & 3--8 & 9--16 & 17--32\\
\midrule
\multicolumn{7}{l}{\emph{strict rule}}\\
Flow-GRPO & relations & 0.93 {\scriptsize(31)} & 0.91 {\scriptsize(21)} & 0.97 {\scriptsize(23)} & 0.95 {\scriptsize(18)} & 1.00 {\scriptsize(7)}\\
Flow-GRPO & other five skills & 0.31 {\scriptsize(16)} & 0.44 {\scriptsize(15)} & 0.82 {\scriptsize(56)} & 0.93 {\scriptsize(91)} & 0.97 {\scriptsize(275)}\\
no KL & relations & 0.92 {\scriptsize(31)} & 0.86 {\scriptsize(21)} & 0.93 {\scriptsize(23)} & 0.97 {\scriptsize(18)} & 0.97 {\scriptsize(7)}\\
no KL & other five skills & 0.55 {\scriptsize(16)} & 0.78 {\scriptsize(15)} & 0.87 {\scriptsize(56)} & 0.90 {\scriptsize(91)} & 0.96 {\scriptsize(275)}\\
\midrule
\multicolumn{7}{l}{\emph{official rule}}\\
Flow-GRPO & relations & 0.96 {\scriptsize(24)} & 0.95 {\scriptsize(12)} & 0.95 {\scriptsize(33)} & 0.99 {\scriptsize(16)} & 0.99 {\scriptsize(15)}\\
Flow-GRPO & other five skills & 0.14 {\scriptsize(8)} & 0.52 {\scriptsize(15)} & 0.75 {\scriptsize(32)} & 0.93 {\scriptsize(58)} & 0.99 {\scriptsize(340)}\\
no KL & relations & 0.91 {\scriptsize(24)} & 0.95 {\scriptsize(12)} & 0.93 {\scriptsize(33)} & 1.00 {\scriptsize(16)} & 0.99 {\scriptsize(15)}\\
no KL & other five skills & 0.52 {\scriptsize(8)} & 0.81 {\scriptsize(15)} & 0.88 {\scriptsize(32)} & 0.90 {\scriptsize(58)} & 0.97 {\scriptsize(340)}\\
\bottomrule
\end{tabular}}
\caption{\textbf{Success rate after RL by the base model's success count} (number of prompts in parentheses). On relations the rate after RL ignores the base rate; on the other skills it tracks it.}\label{tab:buckets}
\end{table}

\section{Training data and its overlap with the evaluation}
\label{sec:data}

\paragraph{Composition.} \Cref{tab:trainset} gives the composition of the three training sets. The released training file has 50{,}000 lines, sampled uniformly, in the proportions 7:5:3:1:1:0 for relations, counting, binding, colours, two objects and single objects~\cite{flowgrpo}. The counting and colour templates admit few combinations, so the 50{,}000 lines contain 33{,}199 distinct prompts, of which 62\% are relation prompts and 0.5\% counting prompts. A skill's share of the training samples and its share of the distinct prompts can therefore differ sixtyfold, and the main text quotes both. Single-object prompts are absent from training altogether: their gain (0.84 to 0.98, $k^\ast{=}4$) is transferred entirely from other skills.

\begin{table*}[!t]\centering\footnotesize\setlength{\tabcolsep}{6pt}
\begin{tabular}{l rrrr rr rr}
\toprule
 & \multicolumn{4}{c}{Flow-GRPO training set} & \multicolumn{2}{c}{60-class split (Run A)} & \multicolumn{2}{c}{relations only (Run B)}\\
\cmidrule(lr){2-5}\cmidrule(lr){6-7}\cmidrule(lr){8-9}
Skill & lines & share & distinct & share & lines & distinct & lines & distinct\\
\midrule
Single object & 0 & 0.0\% & 0 & 0.0\% & 0 & 0 & 0 & 0\\
Two objects & 2{,}941 & 5.9\% & 2{,}941 & 8.9\% & 2{,}941 & 2{,}941 & 0 & 0\\
Counting & 14{,}706 & 29.4\% & 160 & 0.5\% & 14{,}706 & 180 & 0 & 0\\
Colours & 2{,}941 & 5.9\% & 686 & 2.1\% & 2{,}941 & 586 & 0 & 0\\
Position & 20{,}588 & 41.2\% & 20{,}588 & 62.0\% & 20{,}588 & 10{,}777 & 20{,}588 & 10{,}777\\
Attribute binding & 8{,}824 & 17.6\% & 8{,}824 & 26.6\% & 8{,}824 & 8{,}824 & 0 & 0\\
\midrule
\textbf{All} & 50{,}000 & 100\% & 33{,}199 & 100\% & 50{,}000 & 23{,}308 & 20{,}588 & 10{,}777\\
\bottomrule
\end{tabular}
\caption{\textbf{Composition of the training sets.} \emph{lines} are what the sampler draws from; \emph{distinct} counts different prompt strings. The two splits of our controlled runs restrict the templates to 60 of the 80 object classes.}\label{tab:trainset}
\end{table*}

\paragraph{Overlap with the evaluation.} None of the 553 official prompts occurs in the released training set. The second prompt set of \cref{tab:anatomy}b is different: 132 of its 256 prompts occur there, and \cref{tab:split} separates them. Membership makes no consistent difference on two objects, counting or colours. Its 15 zero-coverage relation prompts end at 0.75 and the 32 others at 0.96, the figures quoted in \cref{sec:analysis}. On relations the 32 trained prompts end at 0.92 and the 15 untrained ones at 0.82, a difference that sits in the four untrained zero-coverage prompts (0.48, against 0.84 for the eleven trained ones). The 100 official relation prompts, none of them trained, end at 0.94 with their zero-coverage prompts at 0.93, so being in the training set is not what a relation prompt needs.

For our own runs, 231 of the 553 official prompts occur in the 60-class training set of Run~A, which is why Run~A is evaluated on the other 322; none of the 256 prompts of the second set occurs there. The relation-only training set of Run~B contains 50 of the 100 official relation prompts, all of them with both objects in the training vocabulary. \Cref{sec:controlled} uses this overlap.

\begin{table*}[!t]\centering\footnotesize\setlength{\tabcolsep}{6pt}
\begin{tabular}{l rcccc rcccc}
\toprule
 & \multicolumn{5}{c}{in the training set} & \multicolumn{5}{c}{not in the training set}\\
\cmidrule(lr){2-6}\cmidrule(lr){7-11}
Skill & $N$ & base p@1 & F-G p@1 & resc. & rate$_0$ & $N$ & base p@1 & F-G p@1 & resc. & rate$_0$\\
\midrule
Single object & 0 & -- & -- & -- & -- & 20 & 0.85 & 0.99 & 0/0 & --\\
Two objects & 30 & 0.40 & 0.92 & 0/0 & -- & 18 & 0.41 & 0.97 & 0/0 & --\\
Counting & 31 & 0.54 & 0.97 & 0/0 & -- & 16 & 0.55 & 0.92 & 0/0 & --\\
Colours & 39 & 0.73 & 0.95 & 0/0 & -- & 8 & 0.68 & 0.95 & 0/0 & --\\
Position & 32 & 0.19 & 0.92 & 11/11 & 0.84 & 15 & 0.15 & 0.82 & 4/4 & 0.48\\
Attribute binding & 0 & -- & -- & -- & -- & 47 & 0.30 & 0.85 & 3/3 & 0.68\\
\midrule
\textbf{All} & 132 & 0.48 & 0.94 & 11/11 & 0.84 & 124 & 0.45 & 0.90 & 7/7 & 0.57\\
\bottomrule
\end{tabular}
\caption{\textbf{The second prompt set split by membership in the released training set} (base model against Flow-GRPO, $n{=}32$). Single-object and binding prompts of this set are all untrained.}\label{tab:split}
\end{table*}

\section{Controlled training}
\label{sec:controlled}

\paragraph{Configuration.} \Cref{tab:config} lists the configuration of our runs next to the one released for this task. We changed what a single 24\,GB GPU forces: the group size, the number of groups per epoch and the numerical precision. To fit the memory we keep the frozen transformer weights in half precision, checkpoint activations, split each batch of 8 into micro-batches of 4, and run the T5 text encoder on the CPU with cached embeddings. The configuration asks for two optimizer steps per epoch; the pipeline filters out samples with zero advantage before the update, and 1.9 steps per epoch were applied on average (768 in the first 400 epochs). Checkpoints store the EMA weights. Run~A took 27 GPU-hours. Run~B was trained in five segments of 14 to 26 hours, 108 GPU-hours in total, each segment starting from the previous segment's last checkpoint with a fresh optimizer state.

\begin{table}[H]\centering\small\setlength{\tabcolsep}{4pt}
\resizebox{\linewidth}{!}{\begin{tabular}{lcc}
\toprule
 & Runs A and B & released configuration\\
\midrule
Adapter & \multicolumn{2}{c}{LoRA rank 32 ($\alpha{=}64$), attention projections}\\
Optimizer & \multicolumn{2}{c}{AdamW, lr $3{\times}10^{-4}$, decay $10^{-4}$, clip 1.0}\\
Collection & \multicolumn{2}{c}{SDE, 10 steps, $a{=}0.7$, guidance 4.5, $512^2$}\\
Policy loss & \multicolumn{2}{c}{ratio clip $10^{-4}$, on 9 of the 10 steps}\\
Advantage & \multicolumn{2}{c}{per-prompt mean, global std, clipped at 5}\\
Regularization & \multicolumn{2}{c}{KL weight $\beta{=}0.04$; EMA, decay 0.9}\\
Reward & \multicolumn{2}{c}{graded GenEval reward (\cref{sec:rules})}\\
\midrule
Group size $G$ & 8 & 24\\
Groups per epoch & 8 & 48\\
Images per epoch & 64 & 1{,}152\\
Optimizer steps per epoch & 2 (1.9 applied) & 2\\
Precision & bf16 & fp16\\
GPUs & 1 (24\,GB) & 32\\
\bottomrule
\end{tabular}}
\caption{\textbf{Training configuration.} The upper block is shared; the lower block is what a single GPU changes.}\label{tab:config}
\end{table}

\begin{table*}[!t]\centering\scriptsize\setlength{\tabcolsep}{3.6pt}
\begin{tabular}{l ccccc ccccc ccccc}
\toprule
 & \multicolumn{5}{c}{both objects seen ($N{=}59$)} & \multicolumn{5}{c}{one held out ($N{=}39$)} & \multicolumn{5}{c}{both held out ($N{=}49$)}\\
\cmidrule(lr){2-6}\cmidrule(lr){7-11}\cmidrule(lr){12-16}
 & p@1 & p@32 & $k^\ast$ & resc. & rate$_0$ & p@1 & p@32 & $k^\ast$ & resc. & rate$_0$ & p@1 & p@32 & $k^\ast$ & resc. & rate$_0$\\
\midrule
\sdm base & 0.17 & 0.78 & -- & 0/13 & 0 & 0.14 & 0.59 & -- & 0/16 & 0 & 0.17 & 0.65 & -- & 0/17 & 0\\
Run A, all skills (14k) & 0.19 & 0.75 & 2 & 1/13 & 0.02 & 0.16 & 0.67 & 2 & 0/16 & 0.01 & 0.17 & 0.69 & 2 & 0/17 & 0.01\\
Run B, relations (26k) & 0.24 & 0.73 & 2 & 2/13 & 0.03 & 0.24 & 0.64 & 3 & 0/16 & 0.01 & 0.25 & 0.76 & 2 & 4/17 & 0.04\\
Run B, relations (44k) & 0.30 & 0.86 & 3 & 9/13 & 0.14 & 0.25 & 0.72 & 3 & 2/16 & 0.03 & 0.25 & 0.71 & 2 & 5/17 & 0.06\\
Run B, relations (64k) & 0.40 & 0.88 & 5 & 10/13 & 0.23 & 0.33 & 0.72 & 4 & 2/16 & 0.05 & 0.31 & 0.76 & 3 & 4/17 & 0.09\\
Run B, relations (78k) & 0.53 & 0.98 & 9 & 11/13 & 0.45 & 0.38 & 0.90 & 5 & 8/16 & 0.11 & 0.40 & 0.86 & 5 & 10/17 & 0.14\\
Run B, relations (95k) & 0.54 & 0.93 & 9 & 10/13 & 0.41 & 0.40 & 0.92 & 6 & 8/16 & 0.12 & 0.42 & 0.88 & 5 & 8/17 & 0.13\\
\midrule
Flow-GRPO, released & 0.94 & 1.00 & $>$32 & 13/13 & 0.94 & 0.94 & 1.00 & $>$32 & 16/16 & 0.91 & 0.89 & 1.00 & $>$32 & 16/17 & 0.78\\
\bottomrule
\end{tabular}
\caption{\textbf{Controlled runs on relation prompts, all three groups} (extends \cref{tab:method}; $n{=}32$, image budget in parentheses). \emph{resc.} counts zero-coverage prompts solved at least three times in 32 samples and rate$_0$ is the mean success rate on all of them.}\label{tab:controlled}
\end{table*}

\paragraph{All three groups.} \Cref{tab:controlled} extends \cref{tab:method} with the prompts that have one held-out object and with the ceiling. The one-held-out group behaves like the both-held-out group throughout: at 95k images its zero-coverage prompts are solved at a rate of 0.12, against 0.13 for two held-out objects and 0.41 for none. One unpractised object is enough to delay the unlocking of a prompt. The ceiling itself is noisy at this size: after 26k images it moves by $-0.05$, $+0.05$ and $+0.10$ in the three groups ($+0.03$ pooled), with 13 of the 46 zero-coverage prompts solved at least once and 9 covered prompts dropping to zero, while rate$_0$ stays at or below 0.04. This is why \cref{sec:method} reads coverage from rate$_0$ and from prompts solved at least three times. The released checkpoint, trained on all 80 classes, takes the zero-coverage prompts of the three groups to 0.94, 0.91 and 0.78.

\paragraph{The gap, and what training on a prompt adds.} The left half of \cref{tab:gap} gives the test behind the $p$-values of \cref{sec:method}. The statistic is the difference in mean rate$_0$ between the 13 zero-coverage prompts with both objects seen and the 33 with at least one object held out, compared with 100{,}000 random reassignments of the prompts to the two groups, one-sided. The gap is absent at 26k images ($p{=}0.29$), present from 44k ($p{=}0.005$) and largest at 78k. The test takes prompts as the unit and Run~B as given: it says the gap is not produced by which prompts fell in which group, and it does not say what a second training run would give.

The right half uses the overlap of \cref{sec:data}. Of the 59 relation prompts with both objects seen, 50 are training prompts of Run~B and 9 are not. The two sets start at 0.18 and 0.12 and stand at 0.53 and 0.53 after 78k images, at 0.52 and 0.60 after 95k. Having trained on a prompt confers no measurable advantage over having practised its objects. Nor could it: with about 12k groups over 10{,}777 distinct prompts a training prompt is visited about once, so nothing in Run~B is a best-of-$G$ selection amortized on one prompt.

\begin{table}[H]\centering\small\setlength{\tabcolsep}{3pt}
\resizebox{\linewidth}{!}{\begin{tabular}{l ccc c cc}
\toprule
 & \multicolumn{4}{c}{rate$_0$, zero-coverage prompts} & \multicolumn{2}{c}{p@1, seen objects}\\
\cmidrule(lr){2-5}\cmidrule(lr){6-7}
Images & seen & held out & gap & $p$ & trained & untrained\\
\midrule
0 (base) & 0 & 0 & -- & -- & 0.18 & 0.12\\
26k & 0.03 & 0.02 & 0.01 & 0.294 & 0.25 & 0.19\\
44k & 0.14 & 0.05 & 0.10 & 0.005 & 0.31 & 0.24\\
64k & 0.23 & 0.07 & 0.15 & 0.012 & 0.40 & 0.42\\
78k & 0.45 & 0.13 & 0.33 & $<10^{-4}$ & 0.53 & 0.53\\
95k & 0.41 & 0.13 & 0.28 & 0.0002 & 0.52 & 0.60\\
\bottomrule
\end{tabular}}
\caption{\textbf{Run~B over training} (thousands of images). Left: mean success rate on zero-coverage relation prompts with both objects seen (13 prompts) or at least one held out (33), with a one-sided permutation test over prompts. Right: \passone on the 50 seen-object prompts that are training prompts of Run~B and the 9 that are not.}\label{tab:gap}
\end{table}

\paragraph{Run~A on the official prompts.} \cref{tab:anatomy}b reports Run~A on the second prompt set; \cref{tab:runA} reports it on the 322 official prompts it never saw. Every skill has $k^\ast{=}2$. The pooled ceiling moves from 0.91 to 0.92; single skills move by between $-0.05$ (colours, one prompt of 22) and $+0.06$ (relations, three prompts of 50). Run~A solves 11 of the 30 zero-coverage prompts at least once, eight of them on a single sample; the mean rate over the 30 is 0.02.

\begin{table}[H]\centering\small\setlength{\tabcolsep}{4pt}
\resizebox{\linewidth}{!}{\begin{tabular}{lr cc cc ccc}
\toprule
 & & \multicolumn{2}{c}{base} & \multicolumn{5}{c}{Run A (14k images)}\\
\cmidrule(lr){3-4}\cmidrule(lr){5-9}
Skill & $N$ & p@1 & p@32 & p@1 & p@32 & $k^\ast$ & resc. & rate$_0$\\
\midrule
Single object & 80 & 0.84 & 1.00 & 0.88 & 1.00 & 2 & 0/0 & --\\
Two objects & 49 & 0.48 & 1.00 & 0.54 & 1.00 & 2 & 0/0 & --\\
Counting & 23 & 0.61 & 1.00 & 0.69 & 1.00 & 2 & 0/0 & --\\
Colours & 22 & 0.76 & 1.00 & 0.77 & 0.95 & 2 & 0/0 & --\\
Position & 50 & 0.13 & 0.58 & 0.15 & 0.64 & 2 & 6/21 & 0.01\\
Attribute binding & 98 & 0.37 & 0.91 & 0.44 & 0.93 & 2 & 5/9 & 0.02\\
\midrule
\textbf{All} & 322 & 0.51 & 0.91 & 0.56 & 0.92 & 2 & 11/30 & 0.02\\
\bottomrule
\end{tabular}}
\caption{\textbf{Run~A on the 322 official prompts absent from its training set} ($n{=}32$; columns as in \cref{tab:anatomy}).}\label{tab:runA}
\end{table}

\section{What the KL term holds, prompt by prompt}
\label{sec:kl}

\begin{figure*}[!t]\centering\includegraphics[width=.93\textwidth]{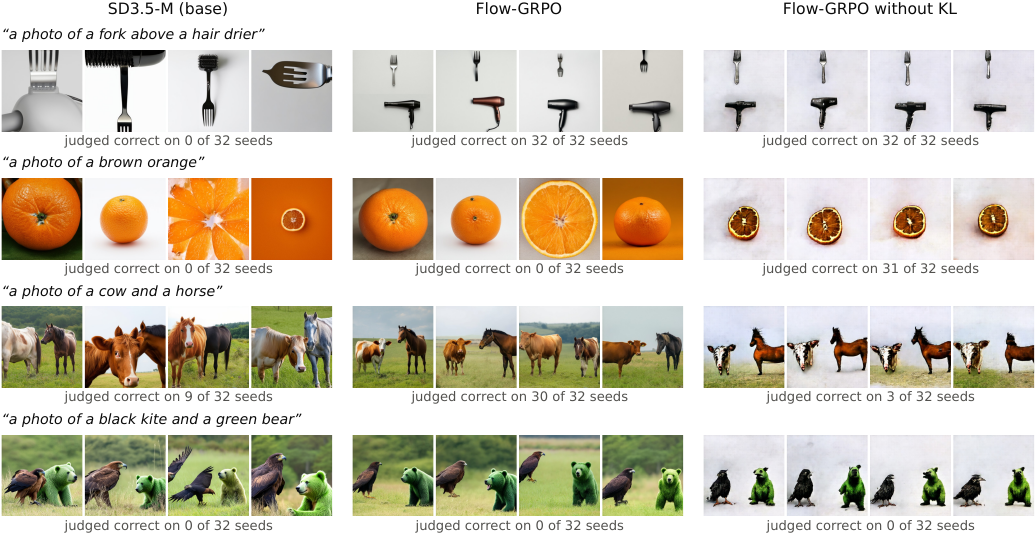}
\caption{\textbf{Four more prompts under the three GenEval models}, four shared seeds each (layout of \cref{fig:qual}). Without KL every image takes the same pale background and one composition.}\label{fig:kl}\end{figure*}

At the level of skills the checkpoints trained with and without KL look alike (\cref{tab:anatomy}a). Prompt by prompt they do not: the mean difference in success rate is $+0.006$ in favour of the checkpoint without KL, but 117 of the 553 prompts differ by more than 0.1, 62 upward and 55 downward. \Cref{tab:nokl} lists the extremes. Four prompts that Flow-GRPO leaves at 3 successes or fewer are solved at least 25 times without KL, among them the two readings discussed in \cref{sec:analysis}. Ten prompts fall more than 0.1 below the base model without KL, against one with KL; nine of the ten had been improved by the regularized checkpoint.

\Cref{fig:kl} shows four more prompts under the three models. ``A fork above a hair drier'' is unlocked with and without KL. ``A brown orange'' stays orange with KL and becomes a dried slice without it. ``A cow and a horse'' is sharpened with KL (9 to 30 of 32) and lost without it (3 of 32), the seeds collapsing onto one composition. ``A black kite and a green bear'' is never judged correct: all three models draw the kite as a bird.

\begin{table}[H]\centering\small\setlength{\tabcolsep}{4pt}
\resizebox{\linewidth}{!}{\begin{tabular}{llrrr}
\toprule
Prompt (``a photo of \ldots'') & skill & base & Flow-GRPO & no KL\\
\midrule
\multicolumn{5}{l}{\emph{held with KL, solved without} (Flow-GRPO $\le 3$ of 32, no KL at least 16 more)}\\
a purple bear & colour & 0 & 0 & 32\\
a brown orange & colour & 0 & 0 & 31\\
a red orange and a purple broccoli & binding & 1 & 0 & 26\\
an orange handbag and a green carrot & binding & 0 & 2 & 25\\
\midrule
\multicolumn{5}{l}{\emph{lost without KL} (more than 0.1 below the base model)}\\
a red umbrella and a blue couch & binding & 24 & 31 & 9\\
a pink skateboard and a black train & binding & 23 & 27 & 14\\
a white sandwich & colour & 9 & 27 & 1\\
a yellow fork & colour & 29 & 32 & 21\\
a blue handbag and a white cell phone & binding & 17 & 29 & 9\\
a yellow suitcase and a brown bus & binding & 18 & 28 & 10\\
a black teddy bear & colour & 31 & 32 & 24\\
a cow and a horse & two obj. & 9 & 30 & 3\\
a toaster and an oven & two obj. & 16 & 23 & 11\\
a blue umbrella & colour & 31 & 31 & 27\\
\midrule
\multicolumn{5}{l}{\emph{lost with KL} (more than 0.1 below the base model)}\\
a person and a snowboard & two obj. & 25 & 21 & 23\\
\bottomrule
\end{tabular}}
\caption{\textbf{Where the two checkpoints part} (successes in 32 samples, strict rule).}\label{tab:nokl}
\end{table}

\section{Text rendering}
\label{sec:ocr}

\begin{figure*}[!t]\centering\includegraphics[width=.93\textwidth]{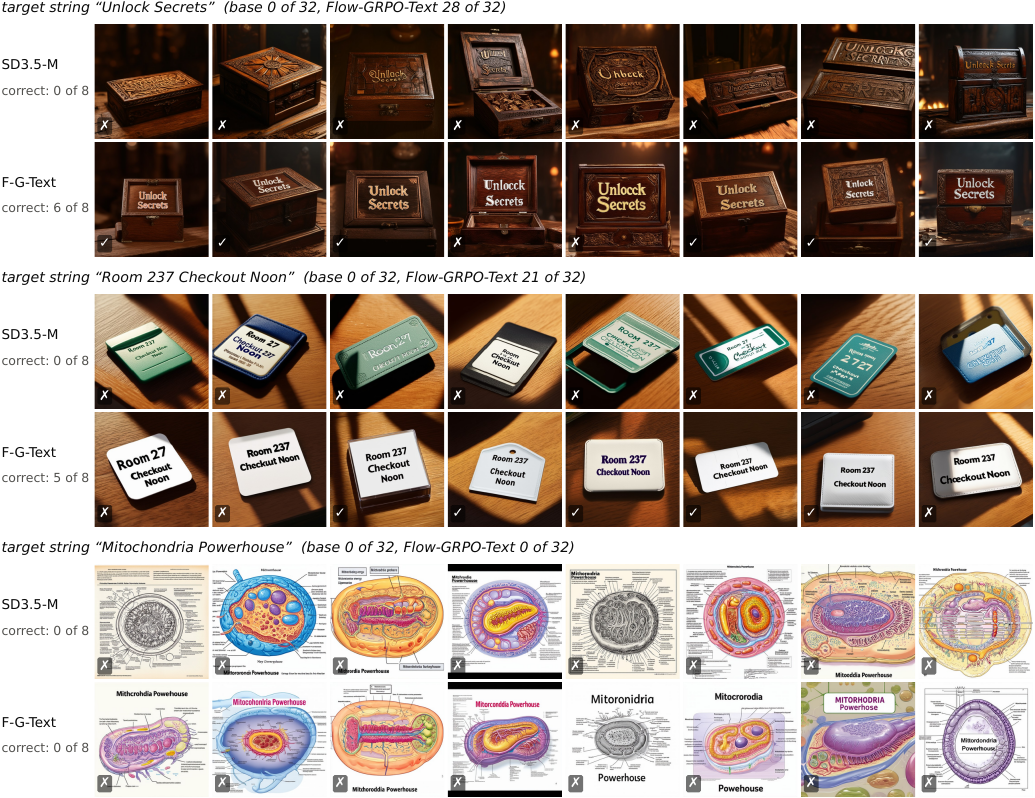}
\caption{\textbf{Text rendering}, eight shared seeds. Two strings the base model never renders in 32 samples and Flow-GRPO-Text renders on most seeds, and one that neither model renders. The mark is the OCR verdict.}\label{fig:ocr}\end{figure*}

\paragraph{Protocol.} The 200 prompts are a random sample of the 1{,}018 test prompts released with Flow-GRPO's text-rendering task; none of them is among its 19{,}653 training prompts. Each asks for a quoted string of 5 to 27 characters, not counting spaces, and we group the prompts by that length: at most 13 (80 prompts), 14 to 16 (58) and 17 or more (62). Images are read with PaddleOCR. The released reward is $1-\min(d,\ell)/\ell$, where $d$ is the edit distance between the target and the recognised text, both lower-cased and stripped of spaces, $\ell$ is the length of the target, and $d{=}0$ if the target occurs as a substring. We call an image correct when the reward is 1, that is, when the target is read back verbatim. The mean reward rises from 0.54 to 0.92 after RL and the strict rate from 0.19 to 0.64.

\paragraph{Mixed, in one table.} \Cref{tab:ocrbuckets} repeats the cut of \cref{tab:buckets} for text. The rate after RL rises with the base success count, from 0.35 to 0.93, as on the five GenEval skills other than relations (0.31 to 0.97, \cref{tab:buckets}) and unlike the flat profile of relations. By this cut text rendering looks sharpened, while its ceiling ($+0.18$) and its 37 rescued strings say otherwise: this is the mixed label of \cref{sec:analysis}. Five strings are rendered by neither model: ``Masters Degree Awarded'', ``Mitochondria Powerhouse'', ``Sea Rover'', ``subsequently'' and ``E=mc\textsuperscript{2}''. \Cref{fig:ocr} shows two rescued strings and one of the five.

\begin{table}[H]\centering\small\setlength{\tabcolsep}{4pt}
\resizebox{\linewidth}{!}{\begin{tabular}{l ccccc}
\toprule
 & \multicolumn{5}{c}{base successes in 32 samples}\\
\cmidrule(lr){2-6}
 & 0 & 1--2 & 3--8 & 9--16 & 17--32\\
\midrule
rate after RL & 0.35 {\scriptsize(42)} & 0.61 {\scriptsize(41)} & 0.65 {\scriptsize(62)} & 0.85 {\scriptsize(40)} & 0.93 {\scriptsize(15)}\\
\bottomrule
\end{tabular}}
\caption{\textbf{Text rendering: success rate after RL by the base model's success count} (200 prompts, $n{=}32$; number of prompts in parentheses).}\label{tab:ocrbuckets}
\end{table}

\section{Diversity among correct samples}
\label{sec:diversity}

\cref{sec:analysis} reports that within-prompt diversity falls by a third after RL and notes that part of the fall is correctness itself, since failed images are varied and the measure includes them. \Cref{tab:diversity} separates the two. The measure is the mean pairwise cosine distance between DINOv2 ViT-B/14 class tokens~\cite{dinov2} of the samples of one prompt, on the 200 prompts and 8 seeds of the 40-step runs. Over all samples it falls from 0.36 to 0.24, a ratio of 0.67. Restricted to correct samples, on the 127 prompts for which both models have at least three, it falls from 0.34 to 0.24, a ratio of 0.71. Removing failures therefore explains little of the fall: the correct images themselves have moved closer together, on every skill. On relations only 7 prompts qualify, because the base model rarely has three correct samples in eight, and the ratio there is 0.69, no longer the lowest (counting 0.58, two objects 0.63): on relations, where the all-sample ratio is 0.57, part of the fall is the removal of failures.

\begin{table}[H]\centering\small\setlength{\tabcolsep}{4pt}
\resizebox{\linewidth}{!}{\begin{tabular}{lr ccc r ccc}
\toprule
 & \multicolumn{4}{c}{all samples} & \multicolumn{4}{c}{correct samples only}\\
\cmidrule(lr){2-5}\cmidrule(lr){6-9}
Skill & $N$ & base & Flow-GRPO & ratio & $N$ & base & Flow-GRPO & ratio\\
\midrule
Single object & 50 & 0.30 & 0.23 & 0.79 & 49 & 0.30 & 0.24 & 0.79\\
Two objects & 34 & 0.38 & 0.24 & 0.63 & 20 & 0.39 & 0.25 & 0.63\\
Counting & 14 & 0.31 & 0.20 & 0.64 & 8 & 0.34 & 0.20 & 0.58\\
Colours & 10 & 0.31 & 0.24 & 0.78 & 9 & 0.30 & 0.22 & 0.76\\
Position & 33 & 0.39 & 0.22 & 0.57 & 7 & 0.29 & 0.20 & 0.69\\
Attribute binding & 59 & 0.39 & 0.26 & 0.66 & 34 & 0.38 & 0.27 & 0.71\\
\midrule
\textbf{All} & 200 & 0.36 & 0.24 & 0.67 & 127 & 0.34 & 0.24 & 0.71\\
\bottomrule
\end{tabular}}
\caption{\textbf{Within-prompt diversity} over all samples and over correct samples only (prompts on which both models have at least three correct samples of eight).}\label{tab:diversity}
\end{table}

\section{The LoRA dial in numbers}
\label{sec:dial}

\Cref{tab:dial} gives the values behind \cref{fig:dial}. All columns use the first 16 seeds, so the ceiling is pass@16 and lower than the pass@32 of \cref{tab:anatomy}. On relations pass@16 goes from 0.60 to 0.96 with half the adapter while \passone is still at 0.59. On the four sharpened skills pass@16 moves by at most 0.05 over the whole range.

\begin{table}[H]\centering\small\setlength{\tabcolsep}{3pt}
\resizebox{\linewidth}{!}{\begin{tabular}{l ccccc ccccc}
\toprule
 & \multicolumn{5}{c}{\passone at LoRA scale $\lambda$} & \multicolumn{5}{c}{pass@16 at LoRA scale $\lambda$}\\
\cmidrule(lr){2-6}\cmidrule(lr){7-11}
Skill & 0 & 0.25 & 0.5 & 0.75 & 1 & 0 & 0.25 & 0.5 & 0.75 & 1\\
\midrule
Single object & 0.86 & 0.90 & 0.95 & 0.97 & 0.99 & 1.00 & 1.00 & 1.00 & 1.00 & 1.00\\
Two objects & 0.50 & 0.67 & 0.82 & 0.90 & 0.95 & 0.95 & 0.99 & 1.00 & 1.00 & 0.99\\
Counting & 0.55 & 0.72 & 0.83 & 0.89 & 0.94 & 0.97 & 0.99 & 0.99 & 1.00 & 1.00\\
Colours & 0.72 & 0.80 & 0.84 & 0.86 & 0.89 & 0.93 & 0.97 & 0.97 & 0.96 & 0.96\\
Position & 0.16 & 0.33 & 0.59 & 0.86 & 0.94 & 0.60 & 0.82 & 0.96 & 1.00 & 1.00\\
Attribute binding & 0.36 & 0.54 & 0.66 & 0.75 & 0.80 & 0.86 & 0.91 & 0.91 & 0.97 & 0.98\\
\midrule
\textbf{All} & 0.51 & 0.65 & 0.77 & 0.87 & 0.92 & 0.88 & 0.94 & 0.97 & 0.99 & 0.99\\
\bottomrule
\end{tabular}}
\caption{\textbf{The LoRA dial} (553 prompts, $n{=}16$): \passone and pass@16 as the Flow-GRPO adapter is scaled by $\lambda$.}\label{tab:dial}
\end{table}

\end{document}